\documentclass{IEEEoj}
\usepackage{cite}
\usepackage{amsmath,amssymb,amsfonts}
\usepackage{graphicx,color}
\usepackage{textcomp}
\usepackage{float}
\usepackage{flushend}
\usepackage{amsmath}
\usepackage{amsthm}
\usepackage{amssymb}

\newtheorem{theorem}{Theorem}
\usepackage{url}
\usepackage{cleveref}
\usepackage{dsfont}
\usepackage{wrapfig}
\usepackage{booktabs}
\usepackage{multicol}
\usepackage{multirow}
\usepackage{tabularx}
\usepackage{diagbox}
\usepackage{graphicx}
\usepackage{subcaption}
\usepackage{array}
\usepackage{siunitx}
\usepackage{algorithm}
\usepackage[noend]{algpseudocode}
\makeatletter
\def\BState{\State\hskip-\ALG@thistlm}
\makeatother
\def\BibTeX{{\rm B\kern-.05em{\sc i\kern-.025em b}\kern-.08em
    T\kern-.1667em\lower.7ex\hbox{E}\kern-.125emX}}
\AtBeginDocument{\definecolor{ojcolor}{cmyk}{0.93,0.59,0.15,0.02}}
\def\OJlogo{\vspace{-14pt}\includegraphics[height=23pt]{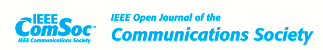}}
\begin{document}

% \receiveddate{XX Month, XXXX}
% \reviseddate{XX Month, XXXX}
% \accepteddate{XX Month, XXXX}
% \publisheddate{XX Month, XXXX}
% \currentdate{XX Month, XXXX}
% \doiinfo{OJCOMS.2022.1234567}

\DeclareRobustCommand*{\IEEEauthorrefmark}[1]{%
  \raisebox{0pt}[0pt][0pt]{\textsuperscript{\footnotesize{#1}}}}

\title{Digital Twin-Enhanced Wireless Indoor Navigation: Achieving Efficient Environment Sensing with Zero-Shot Reinforcement Learning}

\author{Tao Li\authorrefmark{1}, Haozhe Lei\authorrefmark{1}, Hao Guo\authorrefmark{1}\authorrefmark{2}, Mingsheng Yin\authorrefmark{1}, Yaqi Hu\authorrefmark{1}, Quanyan Zhu\authorrefmark{1}, and Sundeep Rangan\authorrefmark{1}}
\affil{Department of Electrical and Computer Engineering, New York University, New York, NY 11201 USA}
\affil{Department of Electrical Engineering, Chalmers University of Technology, Gothenburg, Sweden}
\corresp{CORRESPONDING AUTHOR: Hao Guo (e-mail: hg2891@nyu.edu).}
\authornote{This work was supported, in part, by Swedish Research Council (VR) 2023-00272.}
\markboth{Preparation of Papers for IEEE OPEN JOURNALS}{Author \textit{et al.}}

\begin{abstract}
Millimeter-wave (mmWave) communication is a vital component of future generations of mobile networks, offering not only high data rates but also precise beams, making it ideal for indoor navigation in complex environments. However, the challenges of multipath propagation and noisy signal measurements in indoor spaces complicate the use of mmWave signals for navigation tasks. Traditional physics-based methods, such as following the angle of arrival (AoA), often fall short in complex scenarios, highlighting the need for more sophisticated approaches. Digital twins, as virtual replicas of physical environments, offer a powerful tool for simulating and optimizing mmWave signal propagation in such settings. By creating detailed, physics-based models of real-world spaces, digital twins enable the training of machine learning algorithms in virtual environments, reducing the costs and limitations of physical testing. Despite their advantages, current machine learning models trained in digital twins often overfit specific virtual environments and require costly retraining when applied to new scenarios. In this paper, we propose a Physics-Informed Reinforcement Learning (PIRL) approach that leverages the physical insights provided by digital twins to shape the reinforcement learning (RL) reward function. By integrating physics-based metrics such as signal strength, AoA, and path reflections into the learning process, PIRL enables efficient learning and improved generalization to new environments without retraining. Digital twins play a central role by providing a versatile and detailed simulation environment that informs the RL training process, reducing the computational overhead typically associated with end-to-end RL methods. Our experiments demonstrate that the proposed PIRL, supported by digital twin simulations, outperforms traditional heuristics and standard RL models, achieving zero-shot generalization in unseen environments and offering a cost-effective, scalable solution for wireless indoor navigation.
\end{abstract}

\begin{IEEEkeywords}
Digital Twin, Millimeter-wave (mmWave) Communication, Wireless Indoor Navigation, Reinforcement Learning (RL), Physics-Informed Learning, Zero-shot Generalization.
\end{IEEEkeywords}

\maketitle

\section{INTRODUCTION}
High-frequency transmission in the millimeter-wave (mmWave) bands is a key component of modern fifth-generation (5G) wireless systems, enabling not only massive data rates but also highly accurate positioning and localization capabilities \cite{rangan2014millimeter, dahlman20205g}. The wide bandwidth of mmWave signals, combined with the use of large antenna arrays, allows for fine-grained temporal and angular resolution of signal paths, making mmWave a powerful tool for indoor navigation and simultaneous localization and mapping (SLAM) \cite{shahmansoori20155g, guidi2014millimeter}. Unlike traditional camera-based sensors, mmWave signals provide the added advantage of penetrating beyond line-of-sight, allowing for robust navigation in obstructed and complex indoor environments \cite{ayyalasomayajula2020deep, yin2022millimeter}. However, effectively leveraging mmWave signals for indoor navigation remains challenging. 

In the Wireless Indoor Navigation (WIN) problem \cite{yin2022millimeter}, a mobile robot (or agent) must navigate to a target that broadcasts periodic mmWave signals, while the environment is unknown. Physics-based heuristics, such as following the Angle of Arrival (AoA), provide effective zero-shot generalization in simple settings without requiring training, but they often fall short in complex wireless environments where multipath propagation, including reflections and diffractions, introduces significant challenges \cite{rappaport2015millimeter}. Additionally, the accuracy of such heuristics can be diminished by noisy signal measurements, leading to suboptimal navigation decisions.

To better simulate and evaluate these complex real-world environments, digital twins offer an innovative solution. A digital twin is a virtual replica of a physical system, enabling real-time simulation, optimization, and monitoring across various domains \cite{9854866}. In the context of wireless communications, digital twins are used to model intricate environments and predict how wireless signals behave under different conditions. For indoor navigation, this means a digital twin of a building can simulate how mmWave signals propagate through walls, floors, and other obstacles, providing valuable insights for refining navigation algorithms \cite{9720024}. Digital twins also offer a cost-effective alternative to physical testing, allowing machine learning models to be trained in virtual environments, significantly reducing development costs and improving scalability. Despite these advantages, machine learning models trained in digital twins often suffer from overfitting to specific environments, making them less effective in new settings. Extensive retraining is often required to adapt to different environments, which can be both time-consuming and computationally expensive \cite{9854866}.

In tackling such complex navigation problems besides classic machine learning, deep reinforcement learning (RL) has emerged as a promising end-to-end (e2e) framework, capable of learning policies directly from multimodal input data, including both vision and wireless signals. However, e2e RL methods are data- and computation-intensive, often requiring vast amounts of training data and GPU hours \cite{savva2019habitat}. These models also tend to overfit the training environment, leading to poor generalization when deployed in new, unseen settings \cite{deepmind17forgetting}, and they often require pre-exploration to function effectively in new environments \cite{ayyalasomayajula2020deep}.

\begin{figure*}
    \centering
\includegraphics[width=0.75\textwidth]{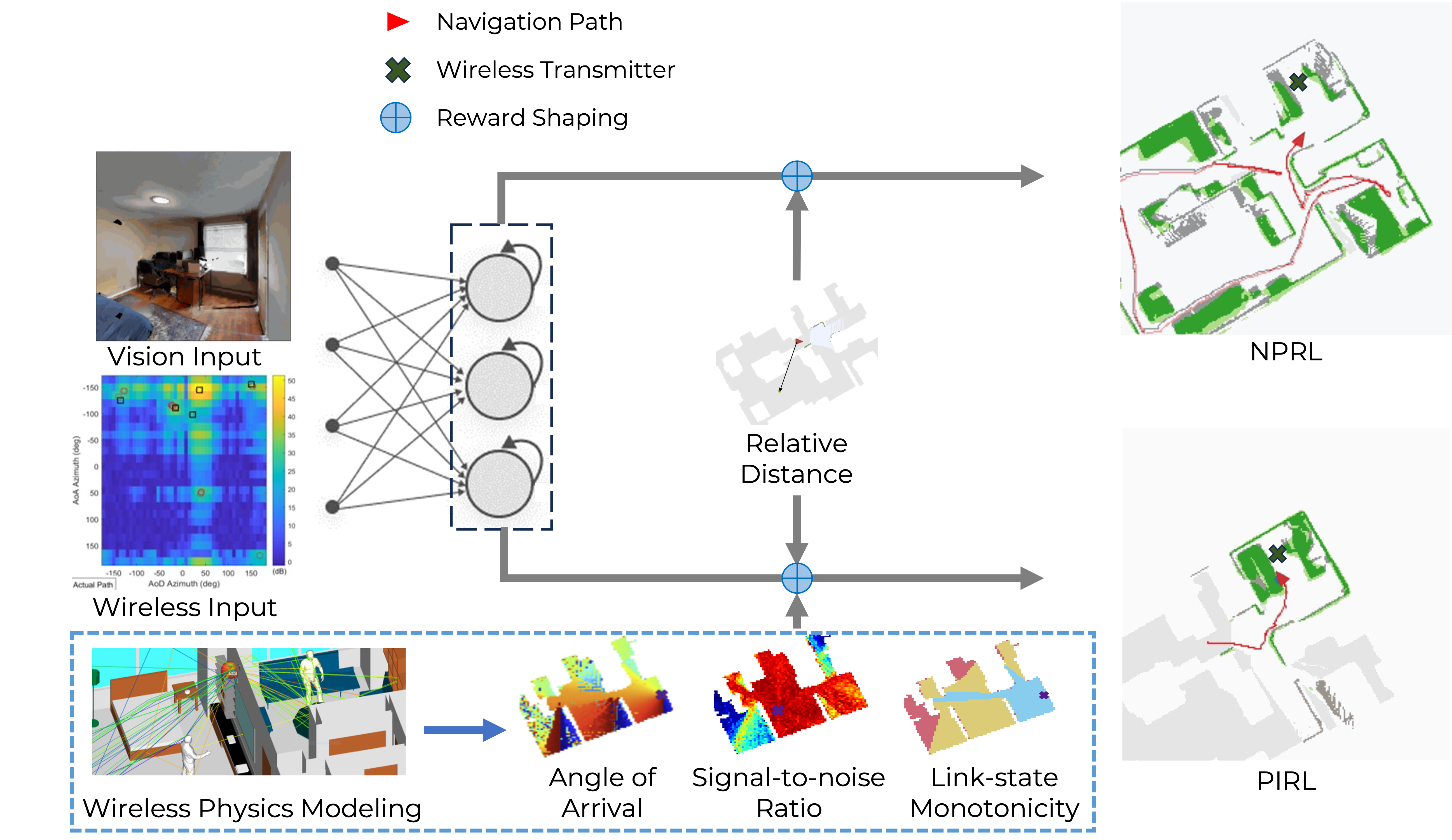}
    \caption{The wireless indoor navigation (WIN) requires the agent to navigate to the wireless transmitter in an unknown environment using multi-modal input, including vision and wireless. The non-physics e2e RL (NPRL), based on relative distance cost, fails to navigate efficiently in unseen scenarios. Trained to utilize physics prior, physics-informed RL (PIRL) acquires zero-shot generalization with interpretable policies.}
    \label{fig:winprob}
\end{figure*}

To overcome these limitations, digital twins can play a pivotal role in making reinforcement learning models more efficient and generalizable. This work proposes a Physics-Informed Reinforcement Learning (PIRL) approach, where physical principles derived from the digital twin environment are incorporated into the RL reward structure. As shown in Fig.~\ref{fig:winprob}, the key idea is to augment the standard e2e RL framework with a reward function shaped by physics-based metrics such as signal strength, AoA, and path reflections. These physically-motivated rewards guide the agent towards actions that align with real-world wireless propagation phenomena, thereby enhancing learning efficiency and improving generalization across different environments. Since these physical principles hold across diverse wireless environments, PIRL enables zero-shot generalization, allowing the trained model to navigate new environments without requiring extensive retraining. As an extension of our previous conference version of the work \cite{10611229}, in this paper, we formulate a more generalized digital twin for the WIN task, and provide deep insights on the algorithm design and extensive new evaluation results.

The contributions of the paper are summarized as follows:  
\begin{itemize}
    \item We design and formulate a Wireless Digital Twin (WDT) framework specifically for the WIN problem, enabling detailed simulation of complex wireless environments.
    \item We propose a novel physics-informed reward shaping approach for RL, simplifying implementation and improving training efficiency by embedding physics-based constraints.
    \item We demonstrate that PIRL reduces the training time and computational overhead compared to vanilla e2e RL, particularly valuable in scenarios where simulating wireless propagation is costly.
    \item Our experiments show that PIRL generalizes well to new environments in a zero-shot manner, outperforming existing heuristic and RL-based methods.
    \item Inspired by recent advances in explainable AI, we perform sensitivity analysis on the learned PIRL model, showing that its actions are interpretable and consistent with the underlying physics principles embedded in the reward function.
\end{itemize}

\section{System Model}
In this section, we first introduce the indoor digital twin platform we developed. Then, we present the problem formulation of the WIN setup.

\subsection{Indoor Digital Twin Formulation}
\label{sec:dt}
\begin{figure*}[!ht]
    \centering
    \begin{subfigure}[t]{0.25\textwidth}
  \centering
  \includegraphics[width=\textwidth]{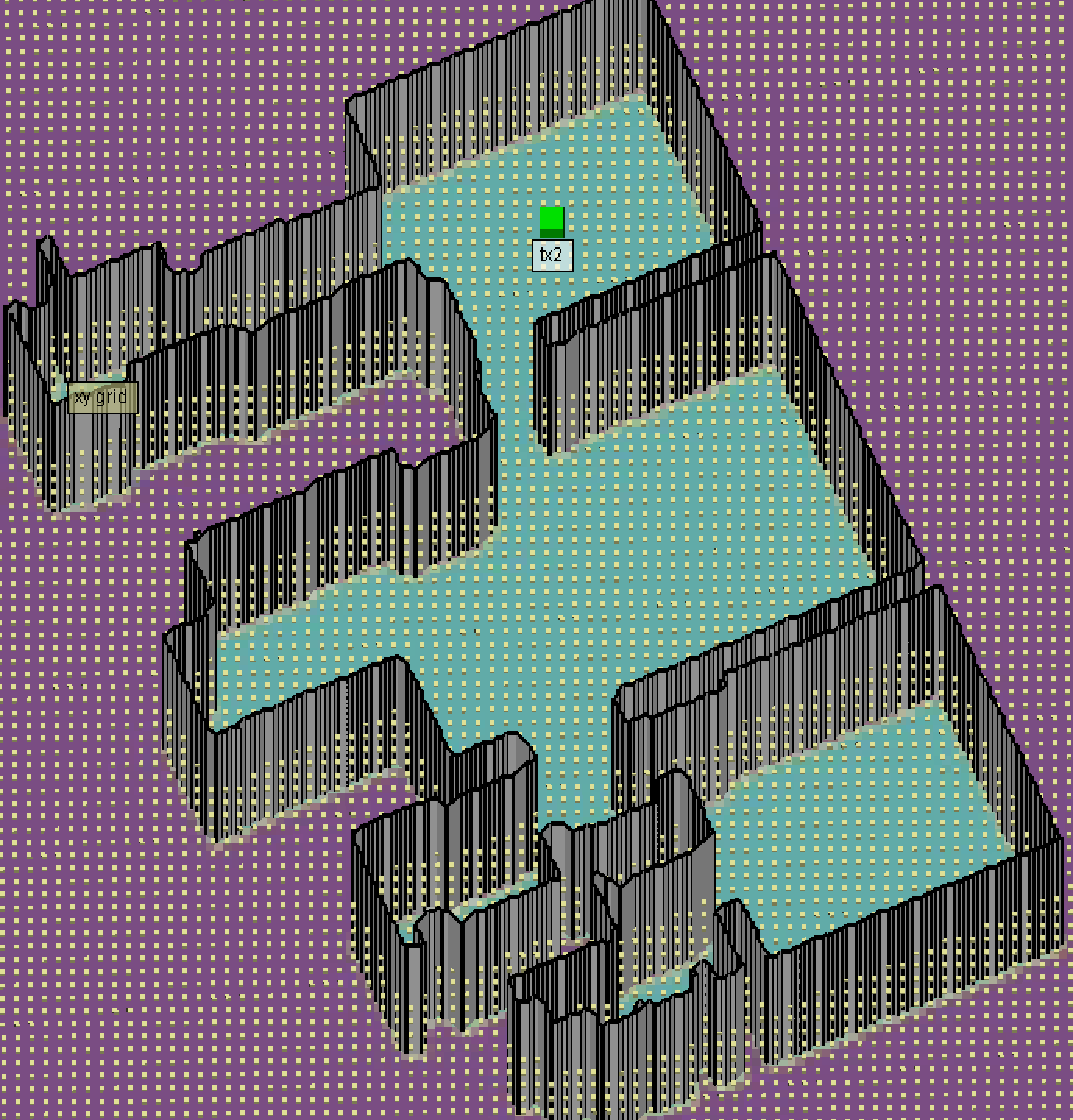}
  \caption{Receiver grid demo.}
  \label{fig:Rx_grid_demo}
\end{subfigure}
\hfill
\begin{subfigure}[t]{0.35\textwidth}
  \centering
  \includegraphics[width=\textwidth]{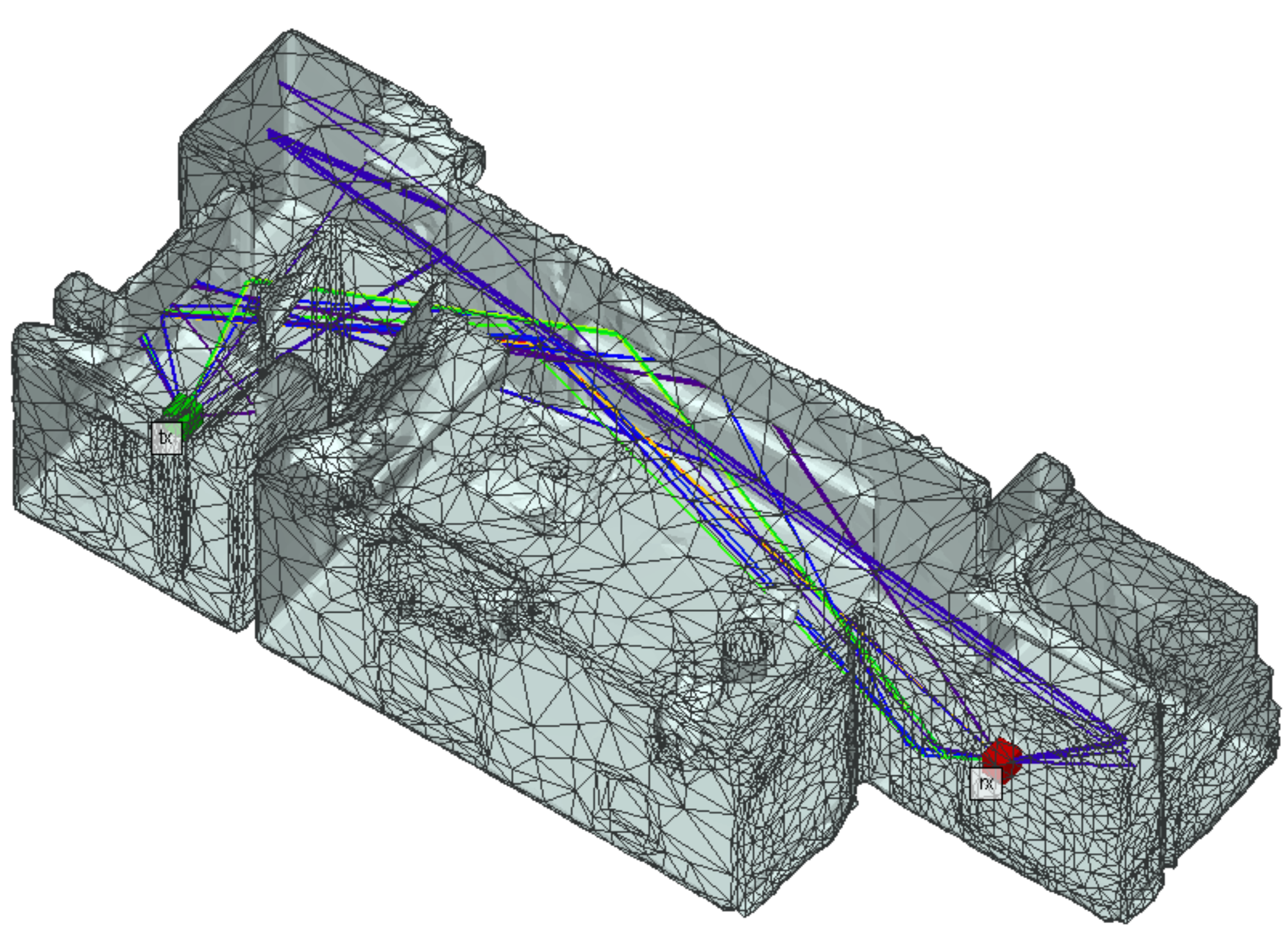}
  \caption{Wireless link in ray tracing demo.}
  \label{fig:ray_tracing_demo}
\end{subfigure}
\hfill
\begin{subfigure}[t]{0.25\textwidth}
  \centering
  \includegraphics[width=\textwidth]{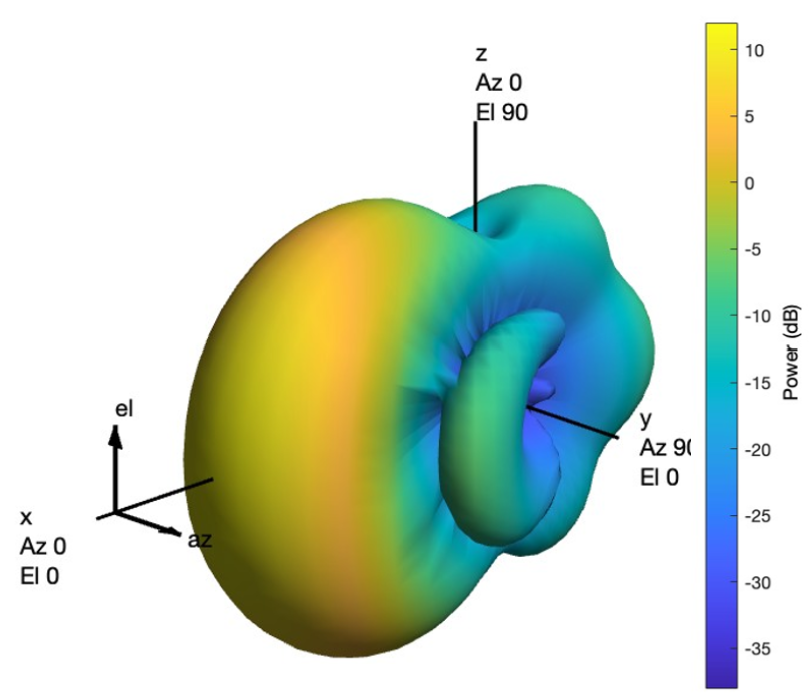}
  \caption{Antenna pattern demo}
  \label{fig:ant_pattern}
\end{subfigure}
\caption{Wireless Channel Simulation Demos}
\end{figure*}

\emph{Wireless Information}:  Since wireless signals and RGB-based images are both multi-dimensional vectors associated with the agent's pose $p$, these two can be treated as vector fields in the considered WIN task. We refer to them as the wireless field $W(p)$ and the vision field $V(p)$, respectively.  For each pose $p$, the wireless field $W(p)$ describes the wireless signal received by the agent at the pose $p$, which includes the AoA and departure (AoD) for five channels. Here, AoA is the direction from which a wireless signal arrives at a receiving antenna, and AoD is the direction in which a wireless signal departs from a transmitting antenna.  Several wireless methods are available to estimate paths from transmitted signals; we use a tensor decomposition method from \cite{yin2022millimeter} reviewed below in \textit{Wireless Digital Twin}.
Mathematically, 
\begin{align}
    W(p)=(g_n, \Omega_n^{rx}, \Omega_n^{tx})_{n=1}^N\in \mathbb{R}^{3\times N},
\end{align}
where $N$ is the maximum number of detected paths, and, for path $n$, $g_n$ denotes its signal-to-noise ratio (SNR),  $\Omega_n^{rx}$ and $\Omega_n^{tx}$ denote the AoA and AoD of the path $n$, respectively.  Following the setup in \cite{yin2022millimeter}, we use the top $N=5$
paths.  

\emph{Wireless Digital Twins}: The genesis of Wireless Digital Twins (WDTs) proposed in this paper relies upon the Gibson model \cite{xia2018gibson}, a remarkable embodiment of real-world indoor reconstruction based on point clouds and RGB-D cameras. The realism of the RGB input $v_t$ in WDT surpasses that of the synthetic SUNCG dataset, earlier utilized in exploration research \cite{song2016ssc}. The simulated wireless field $W(p)$ adheres to the millimeter-wave (mmWave) simulation methodology expounded in \cite{yin2022millimeter}. 

To initiate the wireless data simulation, a mesh discretization of the 2D map with a cell width of \SI{15}{cm} is implemented as presented in Fig.~\ref{fig:Rx_grid_demo}, and wireless signals for each vertex point are generated. The simulation commences with the utilization of ray-tracing software such as Wireless InSite \cite{Remcom} to generate noise-free electromagnetic wave rays, as shown in Fig. \ref{fig:ray_tracing_demo}. However, it is crucial to understand that these ray-tracing wireless propagation paths are not exact representations of real-world wireless channels that a robot could actually receive. That is to say, the robot cannot directly access the ray tracing paths in the real world, indicating a potential deviation between the simulated and real-world environments.

The subsequent phase involves the orchestration of antenna arrays, the induction of noise, and the subsequent disintegration of the channel, enabling the extraction of potential real-world robot-receivable wireless signal paths.

The rendering of high-resolution ray-tracing data to cover the entire map for channel sounding and robot navigation entails the use of a 2D receiver (RX) grid with a \SI{15}{cm} interval, as shown in \ref{fig:Rx_grid_demo}. Each task configuration includes one transmitter (TX) and an RX grid, referred to as a wireless link in wireless communication parlance. The strongest 25 rays out of 250 are chosen for each wireless link to simulate the wireless channel, as validated by numerous experiments \cite{khawaja2017uav, hu2022multi, thrane2020model}.

Then, the subsequent step involves the design of antenna arrays, as shown in Fig.~\ref{fig:ant_pattern}. Leveraging the theoretical foundations laid out in \cite{raghavan2019statistical}, a 1x8 patch microstrip antenna array for the RX and a 2x4 patch microstrip antenna array for the TX are simulated. These arrays enable effective beamforming, a process that involves the manipulation of phase and amplitude of signals from multiple antennas to concentrate signal power in specific directions. To ensure an omnidirectional/360° coverage, three antenna arrays with azimuth angles 0°, 120°, and -120° and 0° elevation are deployed. To facilitate TX and RX detection, a known synchronization signal is transmitted by the TX, sweeping through a sequence of directions from the different TX arrays.

A 3D codebook of the mmWave system is designed following \cite{song2015codebook, xia2020multi} to obtain corresponding AoA and AoD in the channel decomposition post-medium wave. At this point, ray tracing data, antenna patterns, antenna group design, beamforming, and the codebook coalesce to simulate realistic indoor wireless channels. Notably, a loss of 6dB, inclusive of noise figures, is introduced during antenna group design, and additive white Gaussian noise (AWGN) assumed to be independent and identically distributed (i.i.d.) is added across the channel modeling RX antennas. 

With the wireless channel acquired, the next step involves sub-channel (wireless path) estimation via low-rank tensor decomposition \cite{wen2018tensor, zhou2017low}. This yields the wireless data 
\begin{align}
    W(p)=(g_n, \Omega_n^{rx}, \Omega_n^{tx})_{n=1}^5\in \mathbb{R}^{3\times 5},
\end{align}
where $g_n$ denotes the SNR of the $n$-th channel, and $\Omega_n^{rx}$ and $\Omega_n^{tx}$ denote the AoA and AoD of the $n$-th sub-channel.

\begin{figure*}[th]
  \centering
  \includegraphics[width=0.8\linewidth]{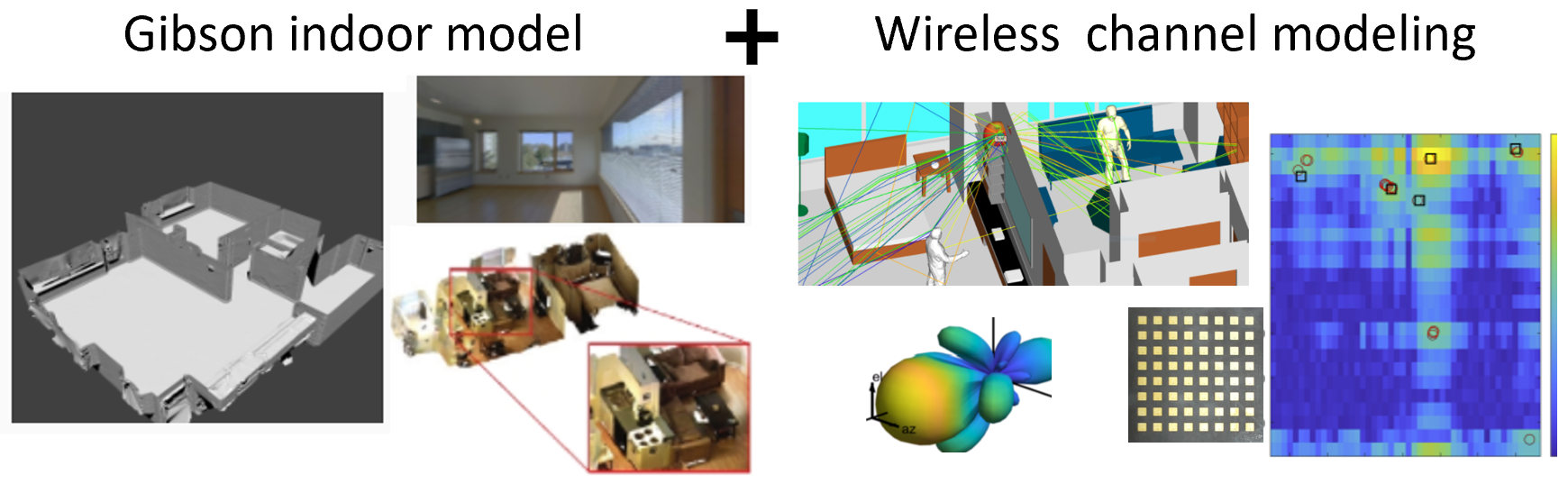}
  \caption{A summary of wireless digital twin (WDT)}
  \label{fig:wdt}
\end{figure*}

Finally, the fusion of the wireless channel data with the Gibson indoor model culminates in the creation of WDTs, meticulously tailored for indoor navigation, as shwon in \ref{fig:wdt}. For additional details regarding the simulation process, readers may refer to the pertinent sections in \cite{yin2022millimeter}.

\subsection{Wireless Indoor Navigation: Task Setup}
\label{sec:win}
With the designed WDT, consider a WIN task setup as studied in \cite{yin2022millimeter}, where a stationary target is positioned at an unknown location in an indoor environment. The target is equipped with a mmWave transmitter that broadcasts wireless signals at regular intervals. Equipped with a mmWave receiver, an RGB camera, and motion sensors, the agent aims to navigate to the target in minimal time. In contrast to the PointGoal task \cite{anderson18evaluate}, WIN does not provide the agent with the target coordinates. The detailed environment setup and the agent's actuation/sensor models are presented below.  

The agent pose is represented by 
\begin{align}
    p=(x,y,\varphi),
\end{align}
where $x, y$ denotes the $xy$-coordinate of the agent measured in meters, and $\varphi$ represents the orientation of the agent in radius (measured counter-clockwise from $x$-axis). Without loss of generality, we assume that the agent starts at $p_1=(0, 0, 0)$.  The agent aims to locate and navigate to the target (the wireless transmitter) denoted by $(x^*, y^*)$.   
We consider a WIN task where the agent operates in the presence of multiple kinds of information feedback
that we denote with a vector 
\begin{align}\label{eq:agent}
    o_t=(m_t, \hat{p}_t, v_t, w_t),
\end{align}
where $t$ is the time step, $m_t$ is an estimate map, $\hat{p}_t = (x_t, y_t)$ is the estimated pose,
$v_t = V(p_t)$ represents visual information, and $w_t = W(p_t)$ represents the wireless information. More details are listed below:
\begin{itemize}
    \item \textbf{Map and pose estimation $m_t$ and $\hat{p}_t$}: The map and pose estimates can come from any SLAM module.  In the simulations below, we will use the state-of-the-art neural SLAM  module proposed in \cite{chaplot2020learning} that provides robustness to the sensor noise during navigation. This SLAM module internally maintains a spatial map $m_t$ and the agent's pose estimate $\hat{p}_t$ (different from the raw sensor reading $\bar{p}_t$) at each time step during the navigation process. The spatial map is represented as  $m_t\in [0,1]^{2\times M\times M}$ is a 2-channel $M\times M$ matrix, where $M\times M$ denotes the map size and each entry corresponds to a cell.  Let $d$ denote the width of the map discretization so each cell is $d \times d$, and the total area is $Md \times Md$, $d=$\,\SI{25}{cm}. Entries of the first channel of $m_t$ denote the probabilities of obstacles within the corresponding cells, while those of the second channel represent the probabilities of the cells being explored \cite{chaplot2020learning}. 
    \item \textbf{Visual information $v_t$}: $V(p)\in \mathbb{R}^{3\times L_1 \times L_2}$ is the 3-channel RGB camera image input at the pose $p$, where $L_1$ and $L_2$ denote the height and the width, respectively. In addition to the wireless sensor and the camera, the agent is also equipped with motion sensors. The sensor readings lead to the estimate of the agent pose $\bar{p}=(\bar{x}, \bar{y}, \bar{\varphi})$, which can be different from the agent's authentic pose $p$. The difference $\varepsilon_{sen}=\bar{p}-p$ is referred to as the sensor noise.  
    \item \textbf{Wireless information $w_t$}:  \begin{align}
        W(p)=(g_n, \Omega_n^{rx}, \Omega_n^{tx})_{n=1}^N\in \mathbb{R}^{3\times N},
        \label{eq:wireless_info}
    \end{align} 
    where $N$ is the maximum number of detected paths along which signals propagate. For the $n$-th path, $g_n$ denotes its signal-to-noise ratio (SNR),  $\Omega_n^{rx}$ and $\Omega_n^{tx}$ denote the angle of arrival (AoA) and departure (AoD), respectively. We consider the top $N=5$ paths with the strongest signal strengths among all paths(see \cite{yin2022millimeter}). 
\end{itemize}

Finally, for agent actions, following \cite{chaplot2020learning}, 
we assume the agent utilizes three default navigation actions, 
\begin{align}\label{eq:actions}
    \mathcal{A}:=\{a_F, a_L, a_R\}.
\end{align}
Here, $a_F=(d, 0, 0)$ denotes the moving-forward command with a travel distance equal to the grid size $d=$\,\SI{25}{cm}; and $a_L=(0,0, -10^\circ)$ and $a_R=(0,0, 10^\circ)$ stand for the control commands: turning left and right by 10 degrees, respectively.  

\subsection{WIN Objective}
Navigating within an unknown environment can be viewed as sequential decision-making using partial observations. The agent's state is given by its authentic pose $p_t$ that remains hidden, and only partial information $o_t$ collected by sensors is available for decision-making at each time step. The state transition as shown in the actuation model presented in Section II-B is Markovian
\begin{align}
    p_{t+1}=p_t + a_t.
\end{align}
Hence, the WIN task is a partially observable Markov decision process (POMDP), with the observation kernel $o_t$ being too complicated to be analytically modeled.

The navigation performance can be measured through a cost function defined as the Euclidean distance (or any distance metric, e.g., geodesic distance) between the current pose and the target position 
\begin{align}
    c_t=\|x_t-x^*\|^2+\|y_t-y^*\|^2.
\end{align}
Denote $\mathcal{H}_t:=\{(o_k, a_k)_{k=1}^{t-1}, o_t\}$ the set of all possible observations up to time $t$, and $\mathcal{H}:=\cup_{t=1}^H \mathcal{H}_t$ the union of all histories, where $H$ denotes the horizon length. The agent's objective in WIN is to find an optimal policy $\pi:\mathcal{H}\rightarrow \mathcal{A}$ such that the cumulative cost $\mathbb{E}_{\pi}[\sum_{t=1}^H c_t]$ is minimized, implying that the agent navigates to the target in minimal time. 

\section{Physics-Informed Reinforcement Learning: Motivations and Reward Shaping with Three Physics Terms}\label{sec:PIRL}
With formulated WIN problem (\ref{eq:agent}) and (\ref{eq:actions}), in this section, we first present the enhanced state-of-the-art method using deep RL and its limitations. Then, we introduce our proposed physics-informed RL (PIRL) with a focus on three specific physics-informed terms: link states, reversibility, and SNR.

\subsection{Classic Deep RL}
The planning algorithms for POMDP \cite{hanna22pomdp} are not suitable for WIN, since the state and the observation space are of high dimensions and continuum, and the observation kernel remains unknown. To create model-free learning-based navigation, one can apply deep reinforcement learning, such as proximal policy optimization (PPO) \cite{schulman2017proximal}, to approximately solve the cost-minimization problem in \eqref{eq:cost_generic}, where the policy $\pi$ is represented by an actor-critic neural network \cite{a3c}, and its model weights are denoted by $\theta\in \mathbb{R}^d$.  
\begin{equation}  \label{eq:cost_generic}
    \min_\theta \mathcal{L}_{\rm RL}(\theta):=\mathbb{E}_{\pi(\theta)} C_{\rm RL}, \quad C_{\rm RL} = \sum_{t=1}^H c_t,
\end{equation}
To address the partial observability in WIN, we incorporate a recurrent module \cite{hausknecht2015deep} into the actor-critic network architecture (see \Cref{sec:policy-struc}). With the recurrent neural network (RNN), the policy $\pi(\theta)$ need not take in all past observations $\{(o_k, a_k)_{k=1}^{t-1}, o_t\}$, and instead, the current partial observation suffices, as RNN can memorize historical input and integrate information feedback across time \cite{hausknecht2015deep}.    We refer to RL with the loss function \eqref{eq:cost_generic} as  \textbf{non-physics-based RL} (NPRL), to differentiate it from the physics-informed RL to be described shortly.

However, as presented in the literature \cite{tao_multiRL, tao_info}, we observe in the initial experiments that when NPRL policies are applied to the WIN problem, they exhibit poor {generalization ability} and {sample efficiency}. For example, the NPRL agent trained for one task (a given map and one target position within the map) even fails to navigate to another target within the same map.  Due to multiple reflections and diffractions of mmWave, the wireless field $W(p)$ changes drastically when the transmitter moves from one location to another, especially when the indoor environment displays complex geometry. Consequently, model weights learned for (overfit) one task are barely relevant to another. In addition to limited generalization, the NPRL agent requires an astronomical amount of samples due to catastrophic forgetting. Since wireless fields vary across different tasks, knowledge of the previously learned task may be abruptly lost as information relevant to the current task is incorporated. Hence, the NPRL agent needs to be re-trained under previous tasks, leading to time-consuming shuffle training \cite{savva2019habitat}.

\subsection{Physics-Informed Reinforcement Learning}
Physics-informed machine learning or RL has emerged as a promising approach to simulate and tackle multiphysics problems in a sample-efficient manner \cite{piml21nature}. The gist is that neural networks can be trained from additional information obtained by enforcing physics laws.  Existing general-purpose strategies of distilling the physics-domain-knowledge into the RL agent include supervised-learning approaches such as imitation learning \cite{hussein18il}, and RL approaches such as offline/batch RL \cite{tao20causal, levine20off} and vanilla RL, i.e., online policy learning \cite{Tao_blackwell}, where the agent repeatedly interact with the digital twin to acquire feedback. This work considers the simple online learning approach based on WDT because we need a fair comparison between our proposed PIRL and other baseline wireless navigation approaches that are based on online RL on sample efficiency and {generalization}.   

Adopting online RL, we thus propose to simply augment the cost with \emph{physically-motivated reward shaping}.
Specifically, the augmented cost function is defined as below. 
\begin{align}
    \mathcal{L}(\theta) :=\mathbb{E}_{\pi(\theta)} \left[ C_{\rm RL} + \lambda_{\rm LS} C_{\rm LS} + 
    \lambda_{\rm AoA} C_{\rm AoA} +  \lambda_{\rm SNR} C_{\rm SNR} \right], \label{eq:pirl}
\end{align}
where the additional terms are motivated by physics principles in WIN: $C_{\rm LS}$, for link-state monotonicity,  $C_{\rm AoA}$ for the angle of arrival direction following, and $C_{\rm SNR}$ for SNR increasing. $\lambda_{\rm LS}$, $\lambda_{\rm AoA}$, and $\lambda_{\rm SNR}$ are weighting constants. In the following, we present these three physics-informed terms in details. 

\begin{figure}[ht]
    \centering
    \begin{subfigure}[t]{0.23\textwidth}
        \centering
        \includegraphics[width=1\textwidth]{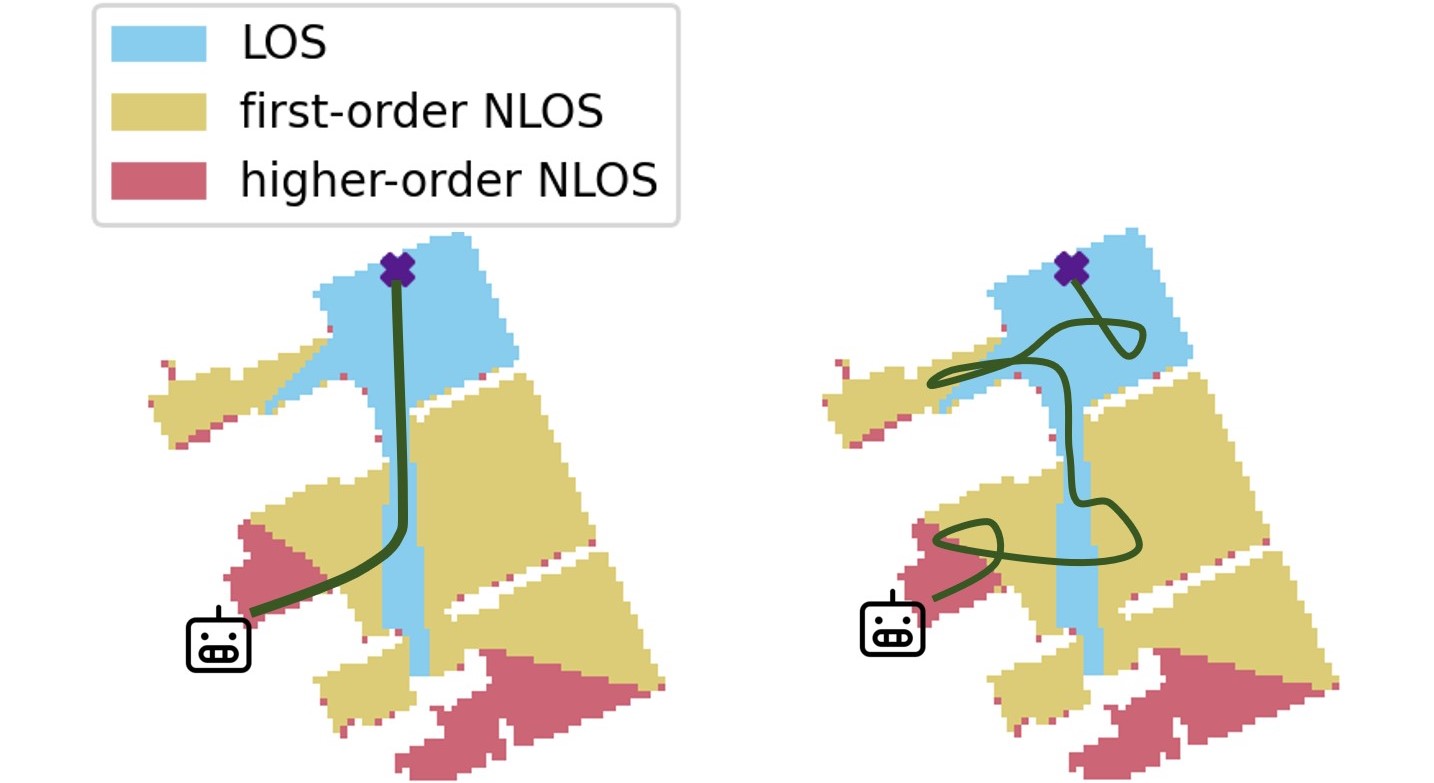}
        \caption{The link state decreases monotonically along the shortest path.}
        \label{fig:principle-1}
    \end{subfigure}
    \hfill
    \begin{subfigure}[t]{0.23\textwidth}
        \centering
        \includegraphics[width=1\textwidth]{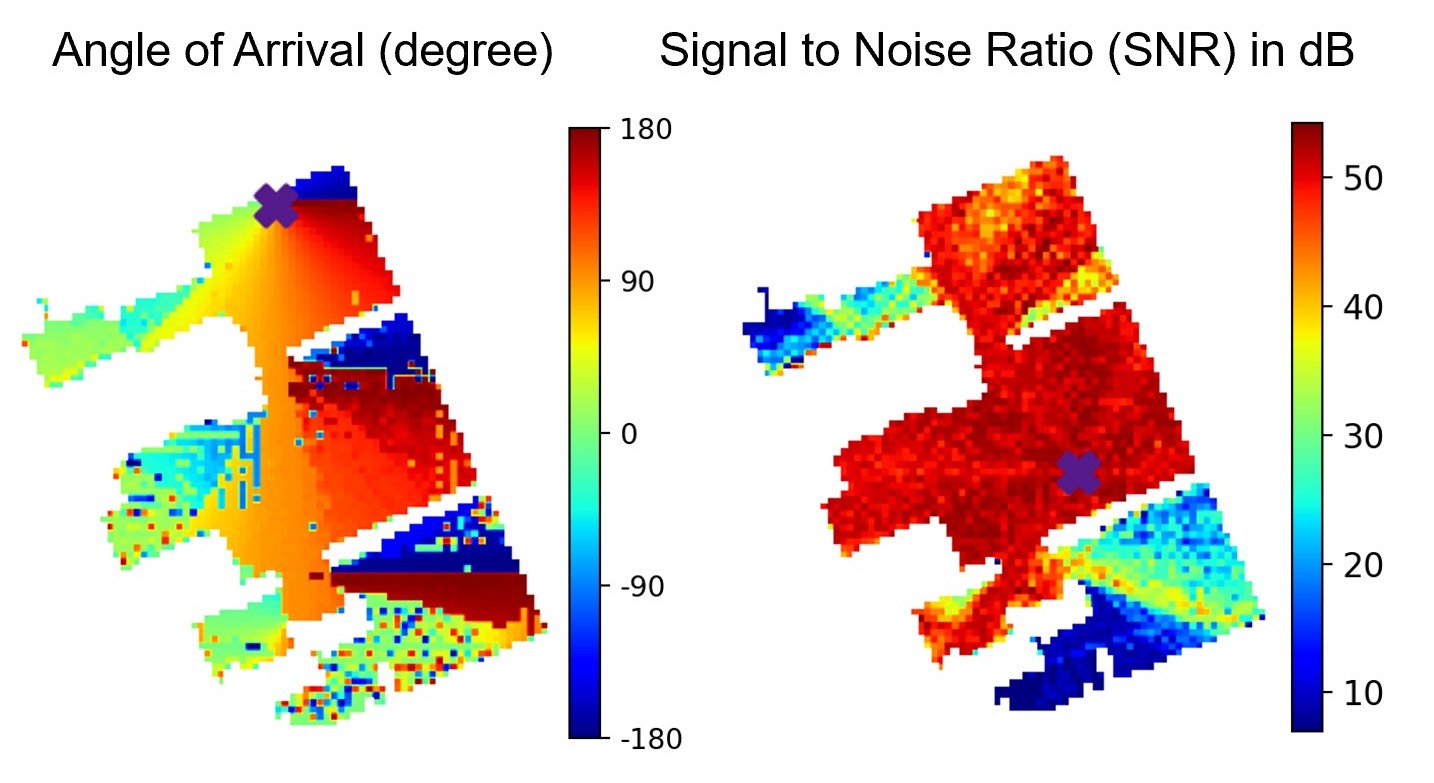}
    \caption{The agent can move reversely along the AoA and explore a high SNR area in NLOS.}
    \label{fig:principle-2-3}
    \end{subfigure}
\caption{The physics principles in WIN.}
\end{figure}

\paragraph{Monotonicity of Link States}  In mmWave applications, link states are of great importance \cite{yin2022millimeter, rangan2014millimeter}, which are primarily categorized into Line-of-Sight (LOS) and Non-Line-of-Sight (NLOS). A location $(x,y)$ (or equivalently a pose $p$) is said to be of \textit{LOS} if there is a wireless signal path wherein electromagnetic waves traverse from the source to the receiver without encountering any hindrances. In contrast, \textit{NLOS} signifies the absence of such a direct visual path. NLOS can further be subdivided into first-order, second-order, third-order, and so forth. First-order NLOS (1-NLOS) implies that at least one electromagnetic wave in the wireless link undergoes a single reflection or diffraction. Likewise, second-order NLOS (2-NLOS) suggests at least one electromagnetic wave undergoing two instances of reflection or diffraction. Similar arguments apply to higher-order NLOS, denoted by $2^+$-NLOS. Define 
\begin{align}
    \ell(p)\in\{0,1,2\}
\end{align}
as the link state of the pose $p$, where the link state evaluation $0$, $1$, and $2$ represent the LOS (0-NLOS), 1-NLOS, and $2^+$-NLOS scenarios, respectively. Note that the link state is a wireless terminology instead of the actual state input to be fed into RL models. Instead, the agent learns to infer the link state from the raw wireless inputs $W(p)$ \cite{yin2022millimeter}.

\Cref{fig:principle-1} presents a distribution map of link state for indoor wireless signals. The purple cross represents the target location. The LOS coverage, by definition, is a connected area, unlike $1$-NLOS, and $2^+$-NLOS coverage. Hence, when the agent enters the LOS area, the shortest path to the target is the straight line connecting the two (see \Cref{fig:principle-1}), which remains within the LOS area. Another important observation is that the LOS area must be bordered by $1$-NLOS, which is then bordered by $2$-NLOS, and so forth. In other words, if the link state increases as the agent navigates, the resulting path cannot be optimal. This observation leads to the following Theorem.

\begin{theorem}\label{theorem1}
A necessary condition for a path to be optimal is that the link state decreases monotonically along the path, which motivates the term 
\begin{align}
    C_{\rm LS} = \sum_t \max\{ 0, {\ell}_t-{\ell}_{t-1}\}.
\end{align}
Mathematically, given navigation path $\vec{p}:=( p_1, \ldots, p_H)$, $p_t$ denotes the pose at time $t$, let $\ell_t=\ell(p_t)$ be the  link state of the pose $p_t$. A necessary condition of $\vec{p}$ being the shortest path is that the link state $\ell$ is non-increasing along the path: $\ell_i\leq \ell_j$, for $0\leq j< i \leq H$.
\end{theorem}
\begin{proof}
Consider a navigation path $\vec{p}:=( p_1, \ldots, p_H)$, $p_t$ denotes the pose at time $t$. Let $\ell_t=\ell(p_t)$ be the corresponding link state of the pose $p_t$. Suppose, for the sake of contradiction, that for the shortest path $\vec{p}$, there exists $0\leq j< i\leq H$ such that $\ell_i> \ell_j$, and we consider two possible cases: 1) $\ell_i=1>0=\ell_j$, and 2) $\ell_i=2>1=\ell_j$. In the first case, when entering the LOS area, the agent shall remain in the LOS, as we discussed earlier. Hence, $\vec{p}$ is not optimal. In the second case, since $\ell$ cannot jump from 2 to 0, there must be some 1-NLOS after $p_i$. Let $k>i$ be the smallest index for which $\ell_k=1$, then connecting $\ell_j$ and $\ell_k$ yields a shorter path, conflicting the optimality. \Cref{fig:principle-1} presents a visualization of the two cases.
\end{proof}

\paragraph{Reversibility of mmWaves} Similar to the principle of reversibility of light, the mmWave follows the same path if the direction of travel is reversed. This reversibility principle leads to a simple yet effective navigation strategy:\textit{ following the angle of arrival (AoA) of the strongest path}, which experiences the least number of reflections.  
Besides, \cite{yin2022millimeter} shows that following the AoA of the strongest path in 1-NLOS cases (NLOS with a single reflection) generally leads to decent navigation since it tends to 
find a route around the obstacle.  However, for 2-NLOS cases (${\ell}_t = 2$), following the AoA may not be a reliable solution,
since it arises from multiple reflections or diffractions. To impose this angle tracking, we add the
term 
\begin{align}
    C_{\rm AoA} = \sum_{t=1}^H |\hat{\Omega}_t-\Omega^{rx}_{1,t}|^2\cdot \mathds{1}_{\{{\ell}_t\neq 2\} }
\end{align}
into \eqref{eq:pirl} where $\hat{\Omega}_t$ is the agent's moving direction derived from the action and $\Omega^{rx}_{1,t}$ is the
AoA of the strongest path included in the wireless information $w_t$.

\paragraph{Navigation in $2^+$-NLOS and the Gradient Field of SNR} Due to reflections, diffractions, and measurement noises, the reversibility principle is less effective in $2^+$-NLOS. Denote by $g(p)=\sum_{i}g_i(p)$ the overall SNR at the pose $p$, or equivalently, the location $(x,y)$. A key observation is that $g$ displays remarkable declines in the transit from the LOS and 1-NLOS to $2^+$-NLOS areas, see \Cref{fig:principle-2-3}. Upon statistically analyzing 21 maps, it is observed that navigating from the 1-NLOS position to the nearest $2^+$-NLOS position leads to an average decline of \SI{25.2}{dB} in SNR. Hence, we propose a navigation heuristic in $2^+$-NLOS scenarios: the agent should move along the reverse direction of the SNR gradient field $-(\nabla_x g, \nabla_y g)$ (finite differences in practice), i.e., toward the direction with the stronger signal strength.  We remark that such a heuristic is less helpful in the LOS and 1-NLOS, where $\nabla g$ is relatively insubstantial: the difference between SNRs of two adjacent mesh vertices is mostly within \SI{3}{dB}.  To encourage the policy to 
increase in SNR, we add the cost 
\begin{align}
    C_{\rm SNR} = \sum_{t=1}^H |\hat{\Omega}_t-\nu_t|^2
\end{align}
where $\nu_t$ denotes the angle between $-\nabla_{x,y} g(p_t)$ and the $x$-axis. In numerical implementations, $\nu_t$ is replaced by the steepest descent direction approximated using finite differences of the mesh points. Consider a discretization of the angle range $\{-180, -170, \ldots, 0, \ldots, 170, 180\}$. For each relative angle from the discrete set, we compute the average of SNR evaluations at all mesh points (where the wireless data is collected, see \Cref{sec:dt}) along the direction. The highest SNR direction is set to be $\nu_t$.

\begin{figure}[ht]
    \centering
    \includegraphics[width=0.4\textwidth]{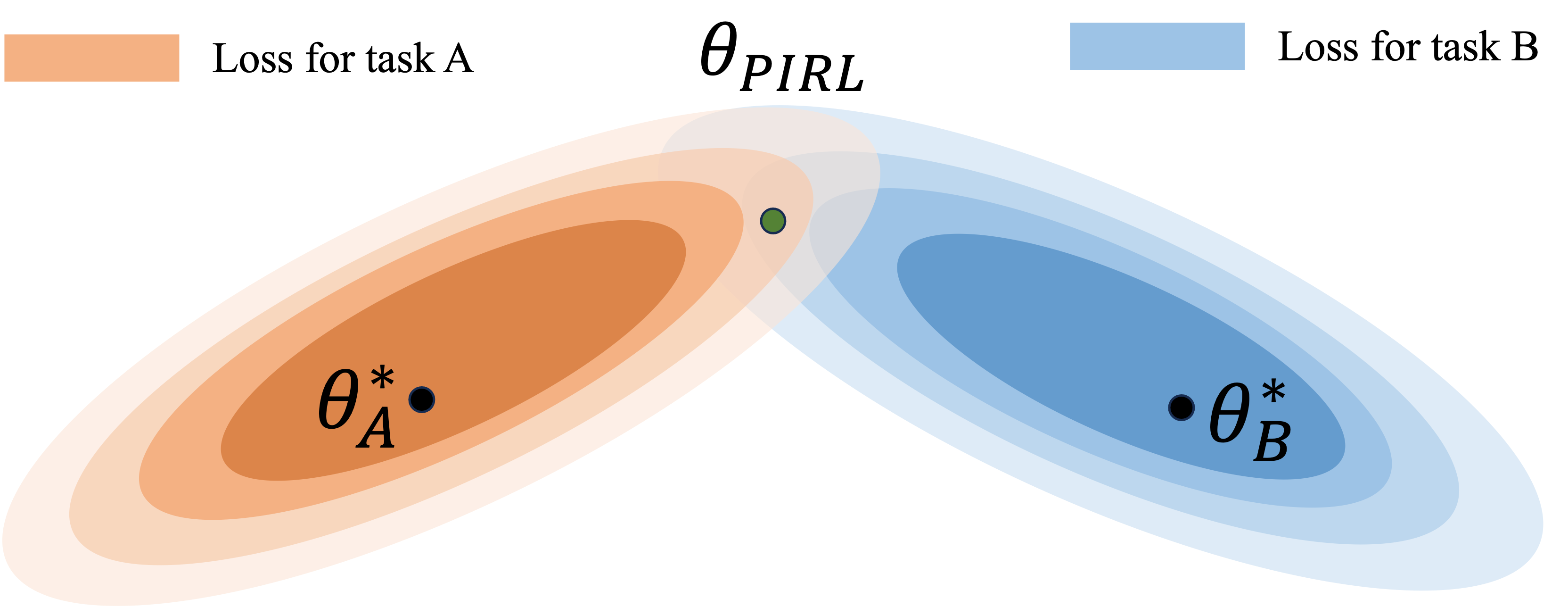}
    \caption{PIRL targets the suboptimal $\theta_{PIRL}$ shared by various tasks, instead of  optimal policies $\theta_A^*$, $\theta_B^*$ for individual tasks.}
    \label{fig:pirl}
\end{figure}

One important observation is that the physics-based reward shaping is not a potential-based transformation \cite{ng99invariance}. To see this, consider a sequence of poses $p_1\rightarrow p_2\rightarrow\cdots\rightarrow p_n\rightarrow \cdots\rightarrow p_1$ such that the agent can travel through them in a cycle, which can incur a net positive cost, e.g., $C_{\rm LS}$ is strictly positive when traversing from LOS to NLOS and then back to LOS. Hence, the policy invariance theorem \cite{ng99invariance} tells that \eqref{eq:pirl} leads to a navigation policy distinct from the shortest path prescribed by \eqref{eq:cost_generic}. For example, following AoA in the 1-NLOS may yield a detour around a corner rather than the shortest path. Even though PIRL is not optimal, it targets suboptimal solution $\theta_{PIRL}$ shared across various tasks (because physics principles are invariant) as shown in \Cref{fig:pirl}. The shared suboptimality alleviates catastrophic forgetting in training and creates zero-shot generalization in testing. 

\section{Proposed PIRL Algorithm and Implementation}

\begin{figure*}[!ht]
     \centering
     \includegraphics[width=1\textwidth]{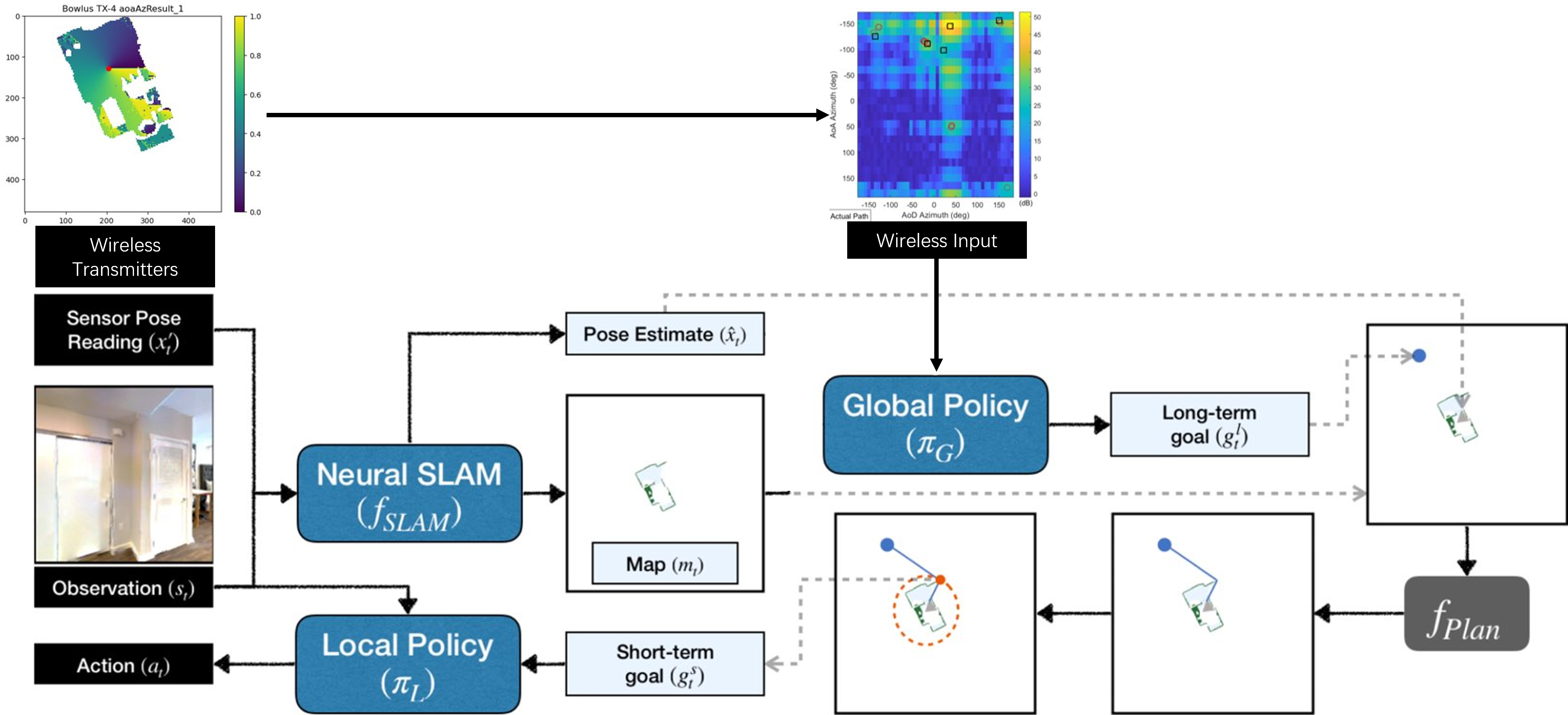}
     \caption{The hierarchical structure of the RL policy. The global policy takes in the wireless input and produces long-term goals (blue dot) fed to the local policy that generates a sequence of navigation actions to an associated short-term goal (red dot). The local policy relies on the active neural-SLAM module \cite{chaplot2020learning} for pose and map estimates that are utilized later by the planner to produce short-term goals.}
     \label{fig:architecture}
 \end{figure*}

To accommodate the heterogeneous information (vision and wireless), we design a hierarchical RL policy inspired by \cite{chaplot2020learning}.  The RL policy consists of two separate neural networks, 
\begin{align}
    \pi(\theta)=(\pi_G(\theta_G), \pi_L(\theta_L)).
\end{align}
Here, $\pi_G$ is a global policy network that sets a long-term goal location, which does not represent the agent's estimate of the target position but rather a waypoint on the navigation path. $\pi_L$ is a local policy that takes in the long-term goal and generates a sequence of navigation actions. A schematic illustration is presented in Fig.~\ref{fig:architecture}, and the pseudocode is summarized in \Cref{alg:training}. The following subsections present the key components of the proposed PIRL algorithm.
\subsection{Overall PIRL Hierarchical Policy Structure}
\label{sec:policy-struc}
Since the wireless information $w_t$ in \eqref{eq:wireless_info} is directly generated by the transmitter, the global policy needs to produce a series of waypoints using such information. Specifically, denote by
\begin{align}
    \alpha_t=\pi_G(w_t\mid\theta_G)
\end{align}
the output from the global policy, which consists of an estimated 
 angle $\hat{\Omega}_t$ and link-state estimates $\hat{\ell}_t$ based on the current wireless input $w_t$. Given the global policy output $\alpha_t$ and the agent current pose estimate $\hat{x}_t, \hat{y}_t$, the long-term goal $p^L_t=(x^L_t, y^L_t)$ can be expressed as
 \begin{align}
     x^L_t=\hat{x}_t+\delta_t \cos{\hat{\Omega}_t},\quad y^L_t=\hat{y}_t+\delta_t\sin{\hat{\Omega}_t},
 \end{align}
where $\delta_t$ is a predicted distance depending on the link state estimate $\hat{\ell}_t$. Also, the predicted distance is given by 
\begin{align}
    \delta_t = \mathds{1}_{\{\hat{\ell}_t = 2\}} \cdot D_b + (1 - \mathds{1}_{\{\hat{\ell}_t = 2\}}) \cdot D_s,
\end{align}
where \(D_b = 7.5\) leading to aggressive exploration and \(D_s = 2.5\) to a conservative one. The intuition behind this setting is as follows: if the agent is in a state of \(2^+\)-NLOS, it prefers to search for the goal aggressively; if not, the agent prefers to move slowly, being more cautious.

Once the global policy determines the long-term goal, a path planner denoted by $f_{plan}$, based on the Fast Marching method \cite{sethian96fmm}, computes the shortest path from the current location to the goal using the spatial map $m_t$ and the pose estimate $\hat{p}_t$ from the SLAM module. The unexplored area is considered a
free space for planning. The output of the planner is a short-term goal
\begin{align}
    p_t^S=f_{plan}(p^L_t, m_t, \hat{p}_t),
\end{align}
which is the farthest point on the path within the grid size $d=0.25$m from the agent. Then, the local policy takes in the path-planning output and the camera images, producing navigation actions 
\begin{align}
    a_t=\pi_L(v_t,p_t^S|\theta_L)
\end{align}
for collision avoidance.

Since the navigation actions correspond to small movements (e.g., turn left/right by $10^\circ$), the agent needs to implement a sequence of actions to move from the current position to the short-term goal before calling the global policy to update the long-term goal. Hence, our PIRL operates the global and local policy at different timescales. Denote by $H_g$ the global decision horizon, indicating the total number of calls to the global policy. At each global time step $t\in \{1,2,\ldots, H_g\}$ (i.e., global policy call), the local policy operates in a local decision horizon denoted by $H_l$: at each local time step $\tau\in \{1, 2, \ldots, H_l\}$, the policy takes in the visual information and executes an action (line 11-14) in Algorithm 1.

We finally conclude the policy network overview by presenting its neural network architecture.  The global policy comprises a recurrent neural network architecture, which includes a linear sequential wireless encoder network with two layers, followed by fully connected layers and a Gated Recurrent Unit (GRU) layer \cite{GRU}. Additionally, there are two distinct layers at the end, referred to as the actor output layer and the critic output layer. The local policy is constructed using a recurrent neural network architecture. It incorporates a pre-trained ResNet18 \cite{resnet18} as the visual encoder, which is followed by fully connected layers and a GRU layer.

\begin{algorithm}[t]
\caption{PIRL  Algorithm}\label{alg:training}
\begin{algorithmic}[1]
  \Require Global policy $\pi_G$, pre-trained Local policy $\pi_L$, and the planner module $f_{PLAN}$;  time horizon $H_g, H_l$
  \State Initialize global policy parameters $\theta_0$;
  \While{not converged}
    \State Reset environment and agent state;
    \State Set global time $t=0$;
    \State Sample initial time $w_t,v_t$ for policies;
    \While{$t < H_g$}
      \State Set local time $\tau = 0$;
      \State Sample action $\alpha_t$ from global policy $\pi_G(w_t|\theta)$;
      \State Compute long-term goal $p_t^L$ using $\alpha_t$;
      \State Compute short-term goal $p_t^S$ using planner $f_{PLAN}$;
      \While{$\tau < H_l$}
        \State Sample action set $a_l$ from local policy $\pi_L(v_t,p_t^S)$;
        \State Execute action $a_l$ and observe next state $v_{\tau+1}$;
        \State Update local time $\tau = \tau + 1$;
      \EndWhile
    \State Update global policy parameter $\theta$ using collected data and the PPO algorithm;
    \State Observe next state $w_{t+1}$;
    \State Update global time $t = t + 1$;
    \EndWhile
  \EndWhile
  \State Output $\theta_t$
\end{algorithmic}
\end{algorithm}

\subsection{Global and Local Policy Training}
\label{sec:global_policy} 
We employ different training processes for the global and local policies since the two policy networks are of different functionalities. For the local policy training, we follow the practice in \cite{chaplot2020learning}, where the local reward is determined by the agent's proximity to the short-term objective and the cross-entropy loss is utilized. The local policy undergoes training via imitation learning, specifically through behavioral cloning, and we refer the reader to \cite{chaplot2020learning} for more details. 

Our contribution mainly lies in global policy training that applies proximal policy optimization (PPO) \cite{schulman2017proximal} with the global reward shaped by physics terms in \Cref{sec:PIRL}. While \eqref{eq:pirl} lays down the general principle for physics-informed reward shaping, the implementation further enforces the monotonicity of link states in the LOS and 1-NLOS, and the resulting global reward function design is as follows.
\begin{equation}
    r_t^g=\left\{\begin{array}{ll}
       \lambda_{\rm LS}C_{\rm LS}\cdot (\zeta_{1} e^{-\zeta_{2} c_t}-\lambda_{\rm AoA}C_{\rm AoA}),  & \hat{\ell}_t\in \{0, 1\} \\
       -\lambda_{\rm SNR}C_{\rm SNR},  &  \hat{\ell}_t\in \{2\},
    \end{array}\right.
\end{equation}
where $\zeta_{1}$ and $\zeta_{2}$ are hyperparameters. The configuration of hyperparameters and weighting constants before each reward term follows the intuition that each component is equally important and shall not dominate or be dominated by other reward components. Note that each reward component falls within intervals of similar ranges. For example, the values for $C_{\rm AoA}$ and $C_{\rm SNR}$ range from 0 to 180. The distance reward, varying across the maps and the target locations, typically assumes a positive value no greater than 200. Since these reward components are of similar ranges, we confine the reward weights to [0.5, 2], over which we utilize a grid search to explore various weight combinations. The detailed hyperparameter setups are deferred to \Cref{sec:exp} 

% In our experiments, we set the values as follows: $\lambda_{\rm LS}=1.1, \lambda_{\rm AoA}=1.0, \lambda_{\rm SNR}=1.2, \zeta_{1}=600$, and $\zeta_{2}=0.1$. 

We are now ready to illustrate the PPO algorithm for global policy training. Denoting 
\begin{align}
    G_t=\sum^{H_g-1}_{k=0}\gamma^k r^g_{t+k+1}
\end{align}
as the discounted future reward starting from $t$, we can derive the state-value function under a global policy $\pi(\theta)$ (also denoted by $\pi_\theta$) as 
\begin{align}
    V_\theta(w) = \mathbb{E}_{\alpha\sim \pi}\left[G_t|w_t\right].
\end{align}
Similarly, we can determine the value function of a (state, action) pair (i.e., $Q$ function) under such policy as 
\begin{align}
    Q_\theta(w,\alpha) = \mathbb{E}_{\alpha\sim \pi}\left[G_t|w_t,  \alpha_t\right].
\end{align}
To measure the performance of an action at a certain state, we can use the advantage function defined as \cite{schulman2017proximal}
\begin{align}
    A_\theta(w,\alpha) = Q_\theta(w,\alpha)-V_\theta(w).
\end{align}

Unlike the vanilla policy gradient method \cite{NIPS1999_464d828b} that optimizes the value function, PPO considers a clipped surrogate objective. Let $\pi_{\theta_{\text {old }}}$ represent the old policy from the last update, and $\pi_{\theta}$ denote the new policy. The probability ratio is denoted as 
\begin{align}
    \mu(\theta) = \frac{\pi_{\theta}(\alpha|w)}{\pi_{\theta_{\text {old }}}(\alpha|w)}.
\end{align}
Additionally, we introduce a small hyperparameter $\epsilon$. To ensure the ratio remains within a certain range, we define the clipping function as 
\begin{align}
    \mbox{clip}(\mu(\theta),1-\epsilon,1+\epsilon).
\end{align}
This function restricts the ratio to be no greater than $1+\epsilon$ and no less than $1-\epsilon$. Therefore, the objective function under this clipping is:
\begin{align}
J^{\mathrm{CLIP}}(\theta) = \mathbb{E}\bigg[&\min \bigg(\mu(\theta) \hat{A}_{\theta_{\text{old}}}(w, \alpha), \\
&\operatorname{clip}(\mu(\theta), 1-\epsilon, 1+\epsilon) \hat{A}_{\theta_{\text{old}}}(w, \alpha)\bigg)\bigg],\nonumber
\end{align}
where $\hat{A}_{\theta_{\text {old}}}(\cdot)$ represents the estimated advantage for the old policy using sample rewards $G_t$. The objective function $J^{\mathrm{CLIP}}(\theta)$ calculates the expectation over the minimum value between two terms: the first term is the product of the ratio and the estimated advantage under the old policy, while the second term is the product of the clipped ratio and the estimated advantage under the old policy. Such an operation addresses the training instability with extremely large parameter updates and big policy ratios.

When implementing PPO on a network architecture with shared parameters for both the policy (actor) and value (critic) functions, the critic is responsible for updating the value function to obtain the estimated advantage function $\hat{A}_{\theta_{\text {old }}}(\cdot)$. On the other hand, the actor serves as our policy model. To promote sufficient exploration in the learning process, an error term, $(V_{\theta}-V_{\text{target}})^2$, and an entropy bonus $H(w,\pi_{\theta}(\cdot|w))$, is introduced for value estimation and exploration encouragement, where $V_{\text{target}}$ represents the discounted cumulative reward associated with a sample trajectory. When a given trajectory ends, target state values are computed as
\begin{align}
    V_t^{\text {target }} =& r_t + \gamma r_{t+1} + \gamma^2 r_{t+2} + \ldots\nonumber\\& + \gamma^{k-1} r_{t+k-1}+\gamma^k V_{\theta_{\text {old }}}\left(w_{t+k}\right), 
\end{align}
where $k$ is the length of trajectory segment. Such segmentation breaks a large sample trajectory into multiple segments, leading to multiple PPO updates along the whole sample trajectory \cite{schulman2017proximal}. In summary, the overall PPO objective function can be written by
\begin{align}
\begin{split}
J^{\mathrm{PPO}}= \mathbb{E}\bigg[&J^{\mathrm{CLIP}}(\theta) - \xi_1(V_{\theta} - V_{\text{target}})^2 + \xi_2H(\pi_{\theta}(\cdot|w))\bigg],
\end{split}
\end{align}
where $\xi_1$ and $\xi_2$ are two hyperparameters. By optimizing the objective function $J^{\mathrm{PPO}}(\theta)$, we obtain the optimal policy $\pi$.

\section{Experiments}
\label{sec:exp}
This section evaluates the proposed PIRL approach for WIN tasks, aiming to answer the following questions. 
\begin{itemize}
    \item \textbf{Sample Efficiency}: does the PIRL take fewer training data than the non-physics-based baseline?
    \item \textbf{Zero-shot Generalization}: can PIRL navigate in unseen wireless environments without fine-tuning? 
    \item \textbf{Interpretablility}: does the PIRL conform to the physics principles, leading to interpretable navigation?
\end{itemize}
We briefly touch upon the training procedure and the experiment setup in the ensuing paragraphs.

\paragraph{Experiment Setup}
\label{app:exp}
The experiment includes 21 different indoor maps (15 for training and 6 for testing) from the Gibson dataset \cite{xia2018gibson} labeled using the first 21 characters in the Latin alphabet (A, B, $\ldots$, U). Table \ref{tab:label_map} presents the label map correspondence, where the left-hand side displays the maps used for training, while the right-hand side displays those for testing. Each map includes ten different target positions labeled using numbers (1,2, $\ldots$, 10). The agent's starting position is fixed for each map regardless of the target position, depending on which, the ten targets for each map are classified into three categories. The first three targets (1-3) are of LOS (i.e., the agent's starting position is within the LOS area), the next three (4-6) belong to $1$-NLOS, and the rest four (7-10) correspond to $2^+$-NLOS scenarios.  For each task (e.g., A1), the maximum number of training episodes is 1000, and the training process terminates if the agent completes the task in more than 6 episodes out of 10 consecutive ones.
\begin{table}[t]
\centering
\caption{Label-Map Correspondence.}
\begin{tabular}{llll|ll}
\toprule
Label & Map Name & Label & Map Name & Label & Map Name \\
\midrule
A & Bowlus & I & Capistrano & P & Woonsocket   \\
B & Arkansaw & J & Delton & Q & Dryville  \\
C & Andrian & K & Bolton & R & Dunmor  \\
D & Anaheim & L & Goffs & S & Hambleton \\
E & Andover & M & Hainesburg & T & Colebrook\\
F & Annawan & N & Kerrtown & U & Hometown\\
G & Azusa & O & Micanopy & & \\
H & Ballou & & & &\\
\bottomrule\label{tab:label_map}
\end{tabular}
\end{table}

During the training phase, the first 15 maps (A-O) with associated 10 task positions are utilized to learn a PIRL policy in sequential order. The training process follows a specific sequence, starting with task A and progressing to A10, followed by training under tasks B1 to B10. Each task consists of 1000 training episodes. This procedure is repeated until the agent has been exposed to all 15 maps with all target positions. The intuition behind this sequential training approach is to gradually increase the complexity of the tasks. It begins with LOS cases, which are relatively simple, then proceeds to 1-NLOS cases, and finally to $2^+$-NLOS cases, which pose a higher level of difficulty. 

% Depending on the task, we use different learning rates on LOS, 1-NLOS, and $2^+$-NLOS which are $1\text{e}^{-4}, 1\text{e}^{-5}, \text{ and } 1\text{e}^{-6}$. For the global horizon $H_g$, we choose $10$, and for the local horizon $H_l$, we choose $20$.

% In contrast to the sequential training of PIRL, we apply rotation training to NPRL, as it suffers from catastrophic forgetting. The rotation training generally follows the task sequence as the sequential training. Yet, after finishing the training on the current task, we randomly select a set of previous tasks to re-train the model before moving to the next task in the sequence to refresh NPRL's ``memory''. The number of re-train tasks is set to be half of the total number of finished tasks. 

% The remaining 6 maps (P to U) are reserved for testing purposes. A total of 20 repeated tests with different random seeds are conducted. For each map, we collect 200 NPLs obtained from these different random seed tests and obtain their average.

\begin{figure*}
 \centering
\includegraphics[width=1\textwidth]{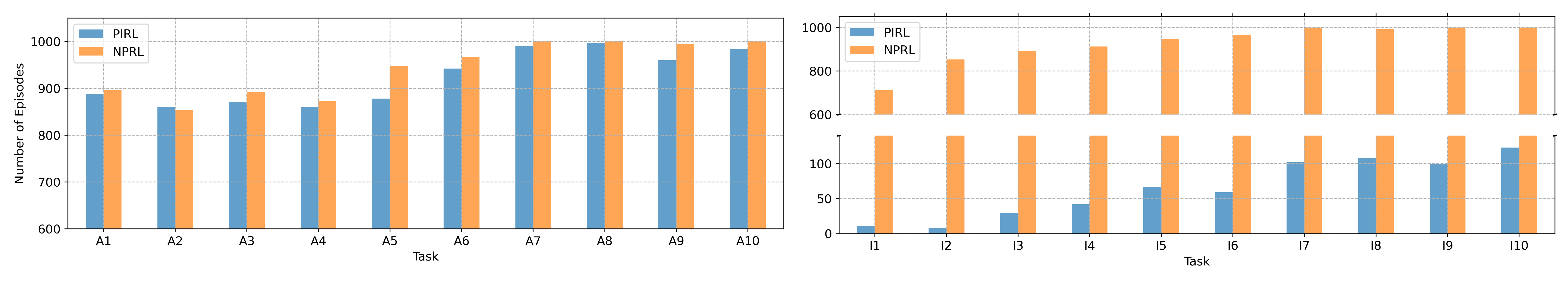}
    \caption{The number of episodes for ten tasks in map A and I. For each map, task number 1-3, 4-6, and 7-10 are tasks of LOS, $1$-NLOS, and $2^+$-NLOS case, respectively. Compared with NPRL, PIRL requires fewer and fewer episodes on each case as the training progresses.  }
    \label{fig:sample}
\end{figure*}

\begin{figure*}
    \centering
    \includegraphics[width=1\textwidth]{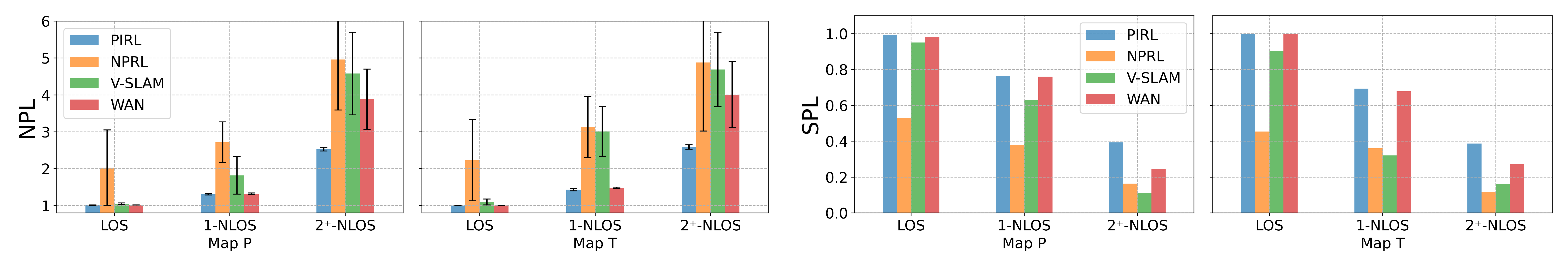}
    \caption{Average NPLs (left) and SPLs (right) returned by navigation policies in the testing. Unlike NPL, SPL uses the inverse of the path length, and hence, the smaller the SPL one returns, the better it is. Since SPL assigns zeros to unsuccessful navigation instances, we do not report its error bar.}
    \label{fig:npl-pt}
\end{figure*}

We consider three baseline navigation algorithms, namely, 
\begin{itemize}
    \item Non-physics-based RL (NPRL): the RL policy is of the same architecture as our proposed PIRL, whereas the reward function is not physics-informed, i.e., only $\mathcal{L}_{RL}$ in \eqref{eq:cost_generic}. 
    \item Wireless-assisted navigation (WAN): this non-RL-based method, put forth in \cite{yin2022millimeter}, relies on a physics-based heuristic that utilizes wireless signals (following AoAs) exclusively within LOS and 1-NLOS scenarios while exploring randomly in $2^+$-NLOS. WAN uses a pre-trained classification model to infer the link state. 
\item Vision-augmented SLAM (V-SLAM), which is a vison-augmented version of the active neural SLAM (AN-SLAM) in \cite{chaplot2020learning}.
\end{itemize}
The first two are primary baselines since our PIRL is a hybrid of the two methodologies. Additionally, to highlight the necessity of leveraging wireless signals in indoor navigation, we consider the third baseline where V-SLAM only takes in RGB image data without wireless inputs. The V-SLAM agent is capable of localizing the target once it falls within the visual (LOS), whereas in the NLOS, V-SLAM reduces to the AN-SLAM, aiming to explore as much space as possible. 

For the first NPRL baseline policy, we consider rotation training to alleviate catastrophic forgetting. The rotation training generally follows the task sequence as the sequential training in PIRL. Yet, after finishing the training on the current task, we randomly select a set of previous tasks to re-train the model before moving to the next task in the sequence to refresh NPRL's ``memory''. The number of re-train tasks is set to be half of the total number of finished tasks. Our experiments use the pre-trained vision model and neural-SLAM module in the other two baselines. We report experimental results based on  20 repeated tests with different random seeds. Moreover, the system parameters, including their detailed descriptions and values in the simulations, are summarized in Table~\ref{tab:hyperparameters}.

\begin{table}
\centering
\caption{Hyperparameters Used in the PIRL Algorithm}
\label{tab:hyperparameters}
\footnotesize
\begin{tabular}{lc}
\toprule
\textbf{Hyperparameter}  & \textbf{Value} \\
\midrule
$\gamma$, discount factor for future rewards & 0.99 \\

$\epsilon$, clipping parameter for PPO & 0.2 \\

$\lambda_{\rm LS}$, weight for link-state cost component & 1.0 \\

$\lambda_{\rm AoA}$, weight for AoA tracking cost component & 0.5 \\

$\lambda_{\rm SNR}$, weight for SNR gradient cost component & 0.3 \\

$\xi_1$, weight for value loss in PPO & 0.5 \\

$\xi_2$, weight for entropy bonus in PPO & 0.01 \\

$\zeta_1$, reward hyperparameter for AoA adherence & 0.7 \\

$\zeta_2$, reward hyperparameter for SNR optimization & 0.3 \\

Learning Rate  & $3 \times 10^{-4}$ \\

Batch Size, Number of samples per training batch & 64 \\

$\alpha$, exploration rate (e.g., epsilon in epsilon-greedy) & 0.1 \\

$\beta$, regularization parameter for policy entropy & 0.01 \\

$H_g$, $H_l$, Global and local decision horizons & 100, 50 \\
\bottomrule
\end{tabular}
\end{table}
\paragraph{Sample Efficiency} We first evaluate the sample efficiency of the PIRL training process by comparing the number of training episodes of PIRL in LOS, 1-NLOS, and $2^+$-NLOS with those of NPRL. The bar plot in \Cref{fig:sample} gives a visualization of the sample efficiency in the training phase on map A (the first map used in the training) and I (midway in the training). In the early stage of the training, no remarkable difference between the two is observed. However, as the training proceeds, PIRL demonstrates a superior sample efficiency on map I, compared with NPRL. This is because the PIRL agent learns to utilize the physics principles that persist across different wireless fields, after being trained on first a few maps. One can see that the PIRL policy already acquires generalization ability to some extent at this point, such that lightweight training would be sufficient for navigating in new environments. In contrast, the NPRL agent, using vanilla end-to-end learning, may be confused when exposed to drastically different wireless fields. Hence, the prior experience does not carry over to the new environment, and NPRL needs to learn almost from scratch.

\paragraph{Generalization} We first highlight that our testing environments (new maps with different target positions) are structurally different
from training cases. {Different room topologies
and wireless source locations create drastically different wireless fields unseen in the training phase, as the reflection and diffraction patterns are distinct across each setup.} We collect the testing performance of three baselines and our PIRL on maps P to U, and report the average results of 20 repeat tests under different random seeds. Since baselines and PIRL use different reward designs, we consider the metric normalized path length (NPL) defined as the ratio of the actual path length (the number of navigation actions) over the shortest path length of the testing task (the minimal number of actions). The closer NPL is to 1, the more efficient the navigation is. The comprehensive comparison is summarized in \Cref{tab:npl}, and \Cref{fig:npl-pt} gives a visualization of NPLs averaged over the LOS task (e.g. P1-3), the 1-NLOS (e.g., P4-6), and the $2^+$-NLOS (e.g., P7-10) on testing maps P and T.  Our PIRL policy generalizes well to these unseen tasks and achieves the smallest NPLs across all three scenarios. In addition to NPL, we also report in \Cref{fig:npl-pt} the Success weighted by (normalized inverse) Path Length (SPL) and in \Cref{tab:success} the success rate, which are customary in the literature \cite{anderson18evaluate}.  
\begin{table*}
\centering
\caption{A comparison of NPLs under 6 testing maps. PIRL achieves impressively efficient navigation in the challenging scenario $2^+$-NLOS, compared with baselines. }
\label{tab:npl}
\fontsize{6.6pt}{8pt}\selectfont
\begin{tabular}{|l|ccc|ccc|ccc|}
\hline
           & \multicolumn{3}{c|}{Map P}                                                                                                                                                                                                            & \multicolumn{3}{c|}{Map Q}                                                                                                                                                                                                            & \multicolumn{3}{c|}{Map R}                                                                                                                                                                                                            \\ \hline
           & \multicolumn{1}{c|}{LOS}                                                        & \multicolumn{1}{c|}{1-NLOS} & \multicolumn{1}{c|}{$2^+$-NLOS} & \multicolumn{1}{c|}{LOS}                                                        & \multicolumn{1}{c|}{1-NLOS} & \multicolumn{1}{c|}{$2^+$-NLOS} & \multicolumn{1}{c|}{LOS}                                                        &\multicolumn{1}{c|}{1-NLOS} & \multicolumn{1}{c|}{$2^+$-NLOS} \\ \hline
PIRL       & \multicolumn{1}{c|}{\begin{tabular}[c]{@{}c@{}}1.01 \\ $\pm$ 0.01\end{tabular}} & \multicolumn{1}{c|}{\begin{tabular}[c]{@{}c@{}}1.31 \\ $\pm$ 0.02\end{tabular}}    & \begin{tabular}[c]{@{}c@{}}2.53 \\ $\pm$ 0.05\end{tabular}     & \multicolumn{1}{c|}{\begin{tabular}[c]{@{}c@{}}1.01 \\ $\pm$ 0.01\end{tabular}} & \multicolumn{1}{c|}{\begin{tabular}[c]{@{}c@{}}1.50 \\ $\pm$ 0.04\end{tabular}}    & \begin{tabular}[c]{@{}c@{}}2.61 \\ $\pm$ 0.05\end{tabular}     & \multicolumn{1}{c|}{\begin{tabular}[c]{@{}c@{}}1.01 \\ $\pm$ 0.00\end{tabular}} & \multicolumn{1}{c|}{\begin{tabular}[c]{@{}c@{}}1.23 \\ $\pm$ 0.03\end{tabular}}    & \begin{tabular}[c]{@{}c@{}}2.55\\ $\pm$ 0.06\end{tabular}      \\ \hline
NPRL        & \multicolumn{1}{c|}{\begin{tabular}[c]{@{}c@{}}2.03 \\ $\pm$ 1.02\end{tabular}} & \multicolumn{1}{c|}{\begin{tabular}[c]{@{}c@{}}2.72 \\ $\pm$ 0.55\end{tabular}}    & \begin{tabular}[c]{@{}c@{}}4.96 \\ $\pm$ 1.37\end{tabular}     & \multicolumn{1}{c|}{\begin{tabular}[c]{@{}c@{}}2.12 \\ $\pm$ 1.00\end{tabular}} & \multicolumn{1}{c|}{\begin{tabular}[c]{@{}c@{}}3.08 \\ $\pm$ 0.68\end{tabular}}    & \begin{tabular}[c]{@{}c@{}}5.00 \\ $\pm$ 1.41\end{tabular}     & \multicolumn{1}{c|}{\begin{tabular}[c]{@{}c@{}}2.28 \\ $\pm$ 1.03\end{tabular}} & \multicolumn{1}{c|}{\begin{tabular}[c]{@{}c@{}}2.49 \\ $\pm$ 0.81\end{tabular}}    & \begin{tabular}[c]{@{}c@{}}4.99 \\ $\pm$ 1.20\end{tabular}     \\ \hline
V-SLAM    & \multicolumn{1}{c|}{\begin{tabular}[c]{@{}c@{}}1.05 \\ $\pm$ 0.02\end{tabular}} & \multicolumn{1}{c|}{\begin{tabular}[c]{@{}c@{}}1.82 \\ $\pm$ 0.51\end{tabular}}    & \begin{tabular}[c]{@{}c@{}}4.58 \\ $\pm$ 1.12\end{tabular}     & \multicolumn{1}{c|}{\begin{tabular}[c]{@{}c@{}}1.11\\ $\pm$ 0.03\end{tabular}}  & \multicolumn{1}{c|}{\begin{tabular}[c]{@{}c@{}}2.89 \\ $\pm$ 0.73\end{tabular}}    & \begin{tabular}[c]{@{}c@{}}4.89\\ $\pm$ 1.00\end{tabular}      & \multicolumn{1}{c|}{\begin{tabular}[c]{@{}c@{}}1.09 \\ $\pm$ 0.03\end{tabular}} & \multicolumn{1}{c|}{\begin{tabular}[c]{@{}c@{}}1.68\\ $\pm$ 0.6\end{tabular}}      & \begin{tabular}[c]{@{}c@{}}4.68 \\ $\pm$ 1.01\end{tabular}     \\ \hline
WAN & \multicolumn{1}{c|}{\begin{tabular}[c]{@{}c@{}}1.02 \\ $\pm$ 0.00\end{tabular}} & \multicolumn{1}{c|}{\begin{tabular}[c]{@{}c@{}}1.32 \\ $\pm$ 0.02\end{tabular}}    & \begin{tabular}[c]{@{}c@{}}3.88 \\ $\pm$ 0.82\end{tabular}     & \multicolumn{1}{c|}{\begin{tabular}[c]{@{}c@{}}1.01 \\ $\pm$ 0.01\end{tabular}} & \multicolumn{1}{c|}{\begin{tabular}[c]{@{}c@{}}1.63 \\ $\pm$ 0.05\end{tabular}}    & \begin{tabular}[c]{@{}c@{}}3.71 \\ $\pm$ 0.71\end{tabular}     & \multicolumn{1}{c|}{\begin{tabular}[c]{@{}c@{}}1.01 \\ $\pm$ 0.00\end{tabular}} & \multicolumn{1}{c|}{\begin{tabular}[c]{@{}c@{}}1.23 \\ $\pm$ 0.02\end{tabular}}    & \begin{tabular}[c]{@{}c@{}}3.97 \\ $\pm$ 0.83\end{tabular}     \\ \hline
           & \multicolumn{3}{c|}{Map S}                                                                                                                                                                                                            & \multicolumn{3}{c|}{Map T}                                                                                                                                                                                                            & \multicolumn{3}{c|}{Map U}                                                                                                                                                                                                            \\ \hline
           & \multicolumn{1}{c|}{LOS}                                                        & \multicolumn{1}{c|}{1-NLOS} & \multicolumn{1}{c|}{$2^+$-NLOS} & \multicolumn{1}{c|}{LOS}                                                        & \multicolumn{1}{c|}{1-NLOS} & \multicolumn{1}{c|}{$2^+$-NLOS} & \multicolumn{1}{c|}{LOS}                                                        & \multicolumn{1}{c|}{1-NLOS} & \multicolumn{1}{c|}{$2^+$-NLOS} \\ \hline
PIRL       & \multicolumn{1}{c|}{\begin{tabular}[c]{@{}c@{}}1.01 \\ $\pm$ 0.01\end{tabular}} & \multicolumn{1}{c|}{\begin{tabular}[c]{@{}c@{}}1.23 \\ $\pm$ 0.01\end{tabular}}    & \begin{tabular}[c]{@{}c@{}}2.82 \\ $\pm$ 0.04\end{tabular}     & \multicolumn{1}{c|}{\begin{tabular}[c]{@{}c@{}}1.00\\ $\pm$ 0.00\end{tabular}}  & \multicolumn{1}{c|}{\begin{tabular}[c]{@{}c@{}}1.43 \\ $\pm$ 0.03\end{tabular}}    & \begin{tabular}[c]{@{}c@{}}2.59\\ $\pm$ 0.06\end{tabular}      & \multicolumn{1}{c|}{\begin{tabular}[c]{@{}c@{}}1.01 \\ $\pm$ 0.01\end{tabular}} & \multicolumn{1}{c|}{\begin{tabular}[c]{@{}c@{}}1.73 \\ $\pm$ 0.03\end{tabular}}    & \begin{tabular}[c]{@{}c@{}}2.46\\ $\pm$ 0.05\end{tabular}      \\ \hline
NPRL        & \multicolumn{1}{c|}{\begin{tabular}[c]{@{}c@{}}2.01 \\ $\pm$ 0.99\end{tabular}} & \multicolumn{1}{c|}{\begin{tabular}[c]{@{}c@{}}2.81 \\ $\pm$ 0.83\end{tabular}}    & \begin{tabular}[c]{@{}c@{}}5.14 \\ $\pm$ 1.21\end{tabular}     & \multicolumn{1}{c|}{\begin{tabular}[c]{@{}c@{}}2.23\\ $\pm$ 1.10\end{tabular}}  & \multicolumn{1}{c|}{\begin{tabular}[c]{@{}c@{}}3.13 \\ $\pm$ 0.83\end{tabular}}    & \begin{tabular}[c]{@{}c@{}}4.88 \\ $\pm$ 1.86\end{tabular}     & \multicolumn{1}{c|}{\begin{tabular}[c]{@{}c@{}}1.90\\ $\pm$ 0.89\end{tabular}}  & \multicolumn{1}{c|}{\begin{tabular}[c]{@{}c@{}}3.25 \\ $\pm$ 0.64\end{tabular}}    & \begin{tabular}[c]{@{}c@{}}4.50\\ $\pm$ 1.05\end{tabular}      \\ \hline
V-SLAM    & \multicolumn{1}{c|}{\begin{tabular}[c]{@{}c@{}}1.04\\ $\pm$ 0.03\end{tabular}}  & \multicolumn{1}{c|}{\begin{tabular}[c]{@{}c@{}}1.99 \\ $\pm$ 0.58\end{tabular}}    & \begin{tabular}[c]{@{}c@{}}4.98 \\ $\pm$ 1.00\end{tabular}     & \multicolumn{1}{c|}{\begin{tabular}[c]{@{}c@{}}1.10 \\ $\pm$ 0.08\end{tabular}} & \multicolumn{1}{c|}{\begin{tabular}[c]{@{}c@{}}3.01 \\ $\pm$ 0.67\end{tabular}}    & \begin{tabular}[c]{@{}c@{}}4.69 \\ $\pm$ 1.01\end{tabular}     & \multicolumn{1}{c|}{\begin{tabular}[c]{@{}c@{}}1.06\\ $\pm$ 0.04\end{tabular}}  & \multicolumn{1}{c|}{\begin{tabular}[c]{@{}c@{}}3.19\\ $\pm$ 0.56\end{tabular}}     & \begin{tabular}[c]{@{}c@{}}4.43\\ $\pm$ 1.00\end{tabular}      \\ \hline
WAN & \multicolumn{1}{c|}{\begin{tabular}[c]{@{}c@{}}1.01 \\ $\pm$ 0.00\end{tabular}} & \multicolumn{1}{c|}{\begin{tabular}[c]{@{}c@{}}1.32 \\ $\pm$ 0.03\end{tabular}}    & \begin{tabular}[c]{@{}c@{}}3.78 \\ $\pm$ 0.90\end{tabular}     & \multicolumn{1}{c|}{\begin{tabular}[c]{@{}c@{}}1.00 \\ $\pm$ 0.00\end{tabular}} & \multicolumn{1}{c|}{\begin{tabular}[c]{@{}c@{}}1.48 \\ $\pm$ 0.02\end{tabular}}    & \begin{tabular}[c]{@{}c@{}}4.01 \\ $\pm$ 0.90\end{tabular}     & \multicolumn{1}{c|}{\begin{tabular}[c]{@{}c@{}}1.01 \\ $\pm$ 0.01\end{tabular}} & \multicolumn{1}{c|}{\begin{tabular}[c]{@{}c@{}}1.74 \\ $\pm$ 0.04\end{tabular}}    & \begin{tabular}[c]{@{}c@{}}3.63 \\ $\pm$ 0.70\end{tabular}     \\ \hline
\end{tabular}
\end{table*}

\begin{table}[]
\centering
\caption{Success Rates in Map T and Map P.}
\label{tab:success}
\begin{tabular}{@{}clllll@{}}
\toprule
\multicolumn{1}{l}{}   &        & PIRL & NPRL & V-SLAM & WAN  \\ \midrule
\multirow{3}{*}{Map T} & LOS    & 1    & 1    & 1      & 1    \\
                       & 1 NLOS & 1    & 1    & 1      & 1    \\
                       & 2+NLOS & 1    & 0.4  & 0.65   & 0.9  \\ \midrule
\multirow{3}{*}{Map P} & LOS    & 1    & 1    & 1      & 1    \\
                       & 1 NLOS & 1    & 1    & 1      & 1    \\
                       & 2+NLOS & 1    & 0.45 & 0.4    & 0.85 \\ \bottomrule
\end{tabular}
\end{table}

\paragraph{Interpretable Navigation}
\begin{figure}
    \centering
    \begin{subfigure}[t]{0.23\textwidth}
    \centering
        \includegraphics[width=1\textwidth]{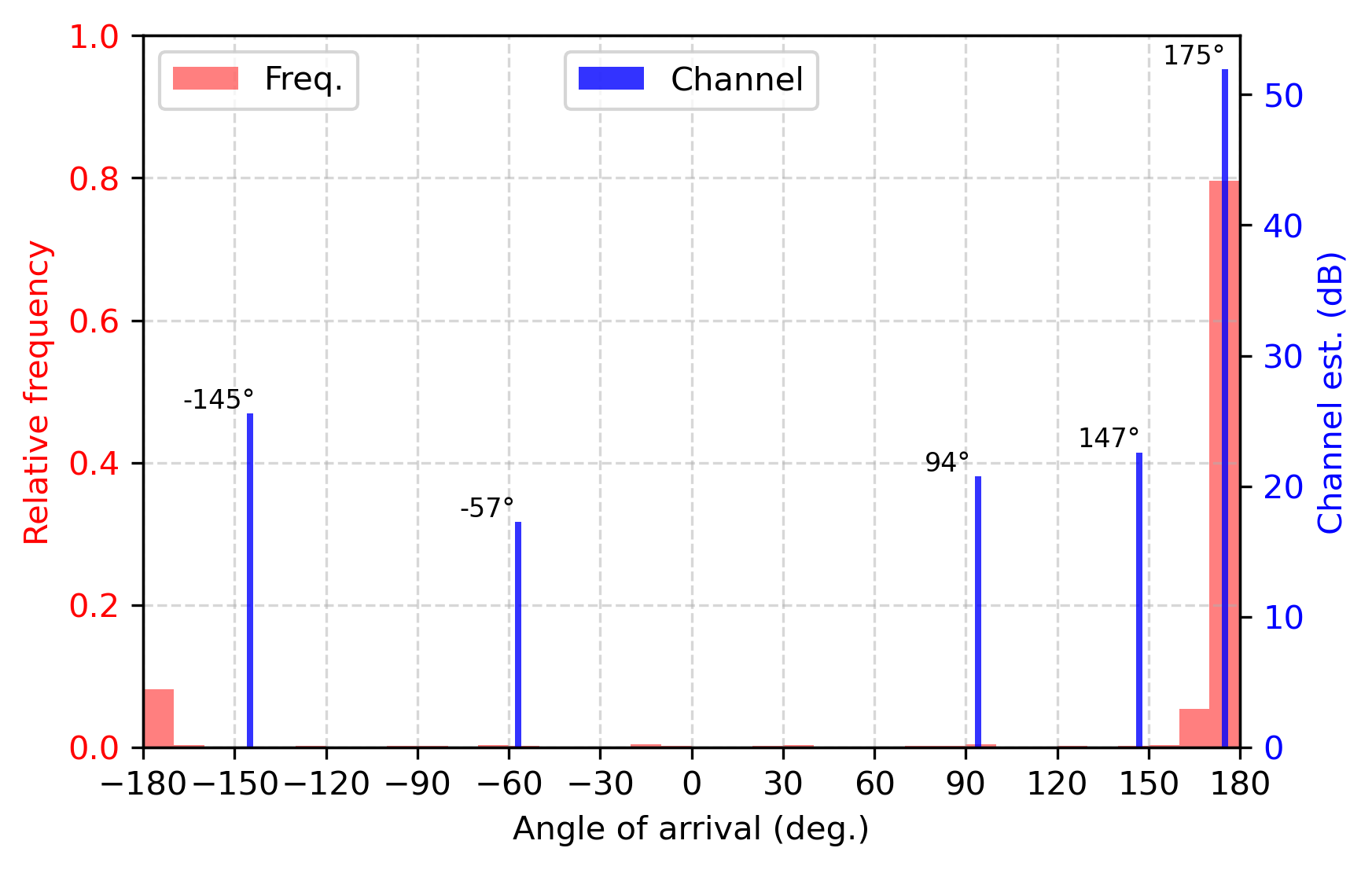}
        \caption{In LOS, PIRL follows the AoA of the first channel (strongest path).}
        \label{fig:aoa-ex}
    \end{subfigure}
    \hfill
    \begin{subfigure}[t]{0.23\textwidth}
    \centering
        \includegraphics[width=1\textwidth]{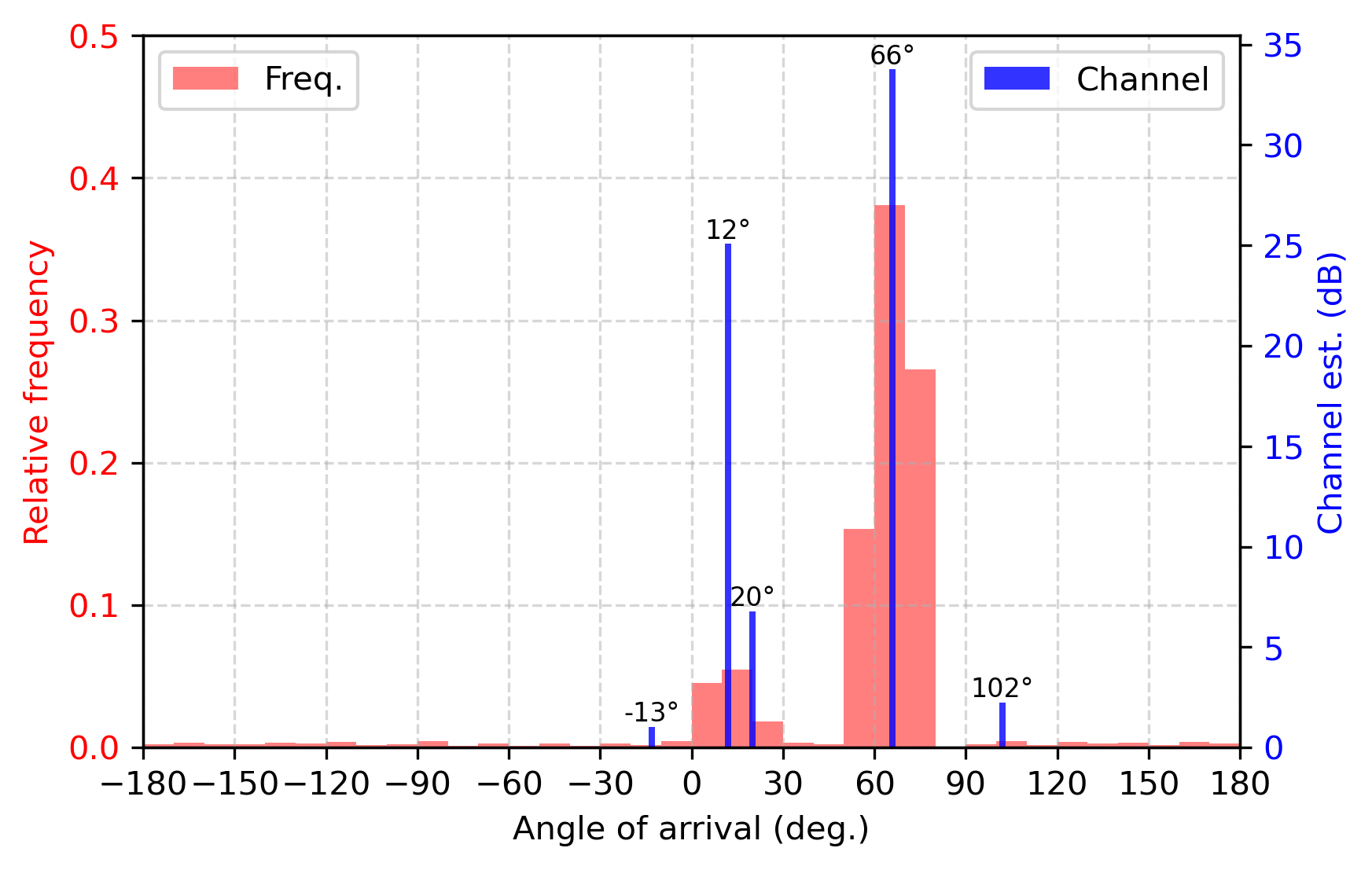}
        \caption{In 1-NLOS, PIRL follows the AoA of the strongest channel.}
        \label{fig:aoa-nlos-ex}
    \end{subfigure}
    \hfill
    \begin{subfigure}[t]{0.23\textwidth}
    \centering
        \includegraphics[width=1\textwidth]{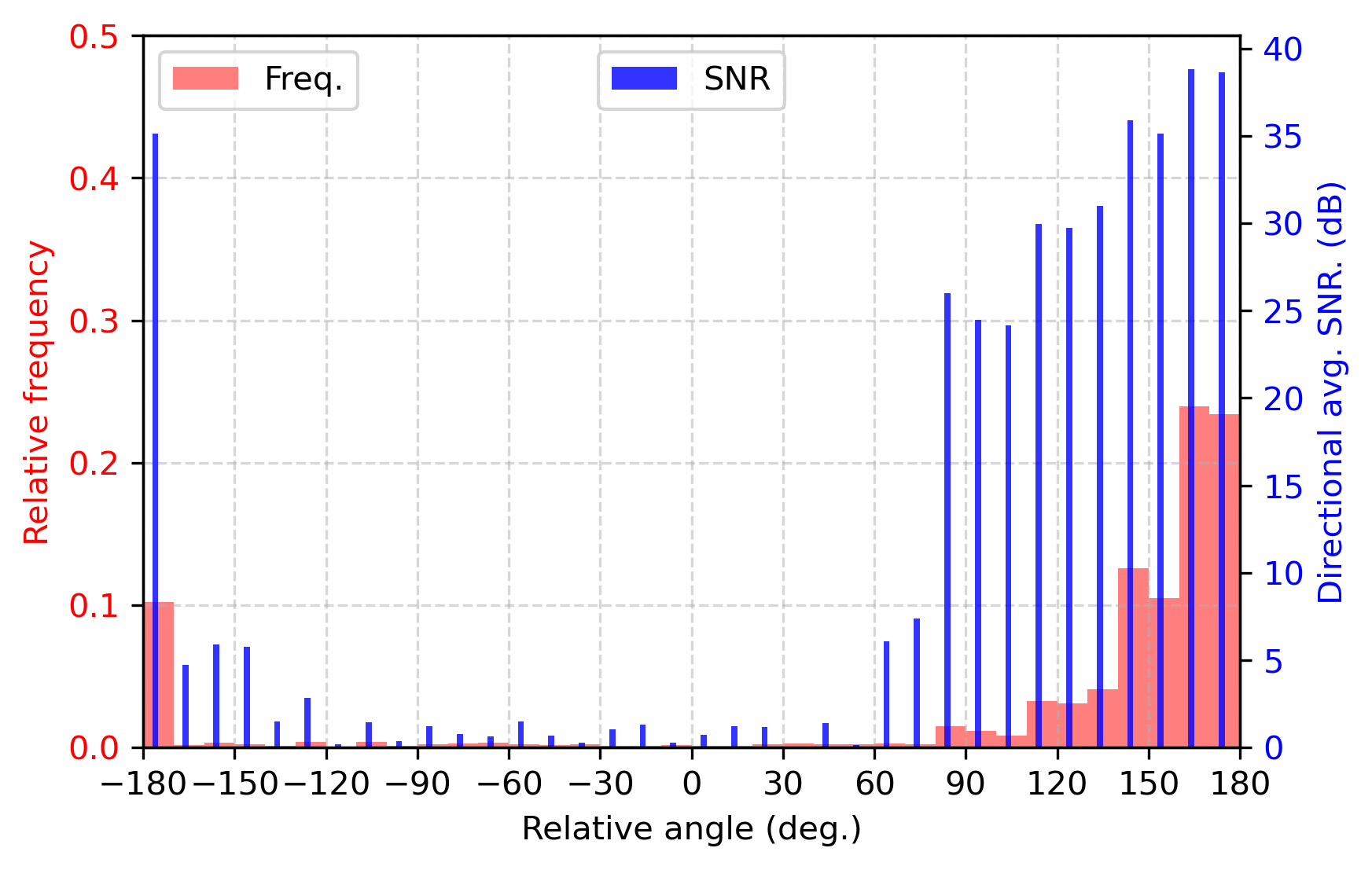}
        \caption{In  $2^+$-NLOS, PIRL aims at high-SNR directions.}
        \label{fig:snr-ex}
    \end{subfigure}
    \hfill
    \begin{subfigure}[t]{0.23\textwidth}
    \centering
        \includegraphics[width=\textwidth]{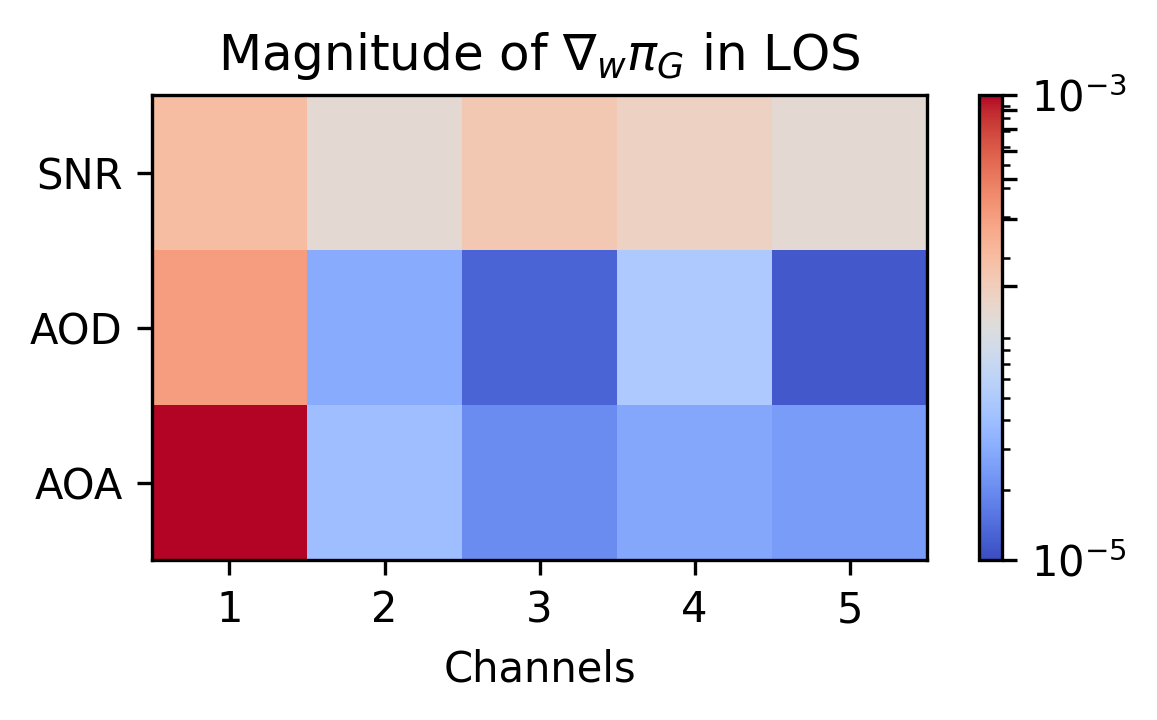}
        \caption{In LOS, the PIRL policy output is mostly sensitive to the AoA of the first channel.}
        \label{fig:grad-aoa}
    \end{subfigure}
    \hfill
    \begin{subfigure}[t]{0.23\textwidth}
    \centering
        \includegraphics[width=\textwidth]{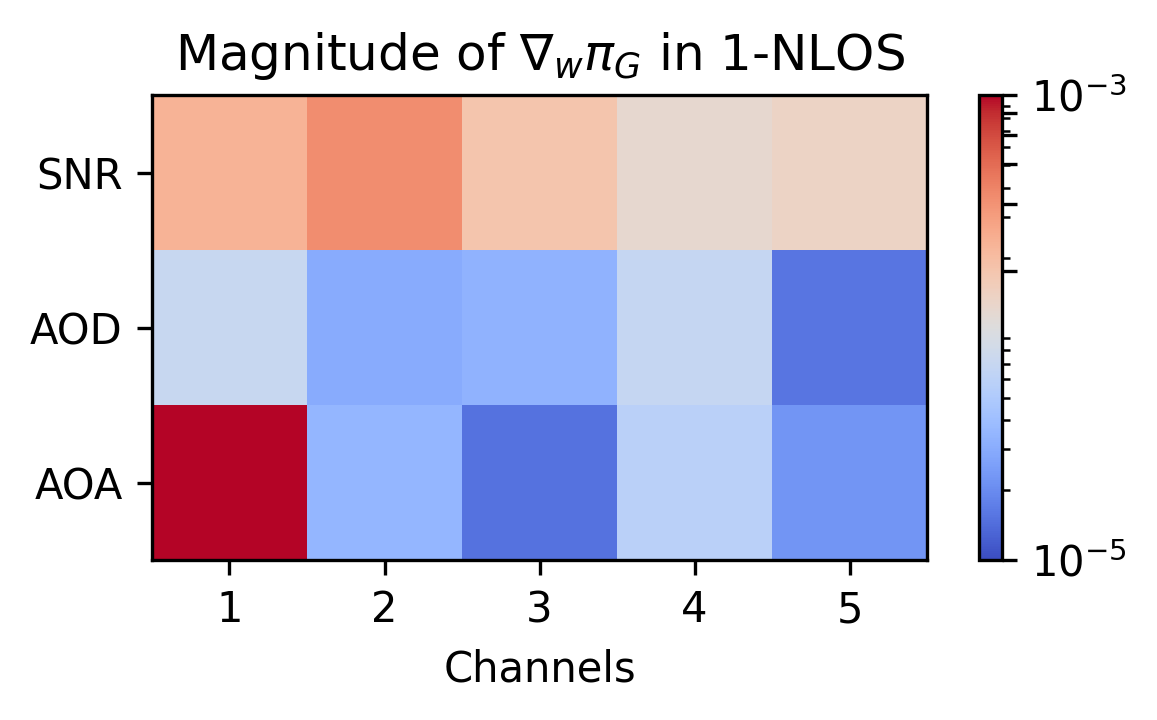}
        \caption{In 1-NLOS, the PIRL policy output is mostly sensitive to the AoA of the first channel.}
        \label{fig:grad-aoa-nlos}
    \end{subfigure}
    \hfill
    \begin{subfigure}[t]{0.23\textwidth}
    \centering
        \includegraphics[width=\textwidth]{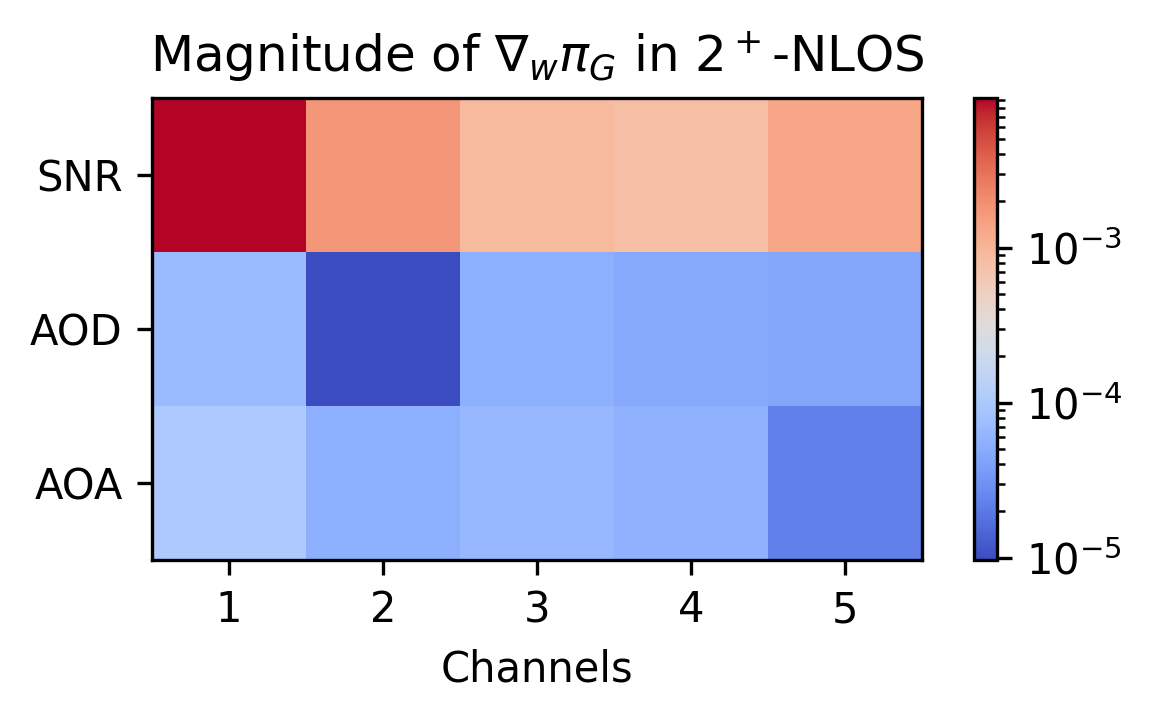}
        \caption{In $2^+$-NLOS, the PIRL policy output is mostly sensitive to the SNR of the first channel.}
        \label{fig:grad-snr-ex}
    \end{subfigure}
    \caption{The interpretability experiments on the reversibility principle and the SNR heuristic.}
    \label{fig:ex-pirl}
\end{figure}
We provide empirical evidence that the PIRL leverages the principles stated in \Cref{sec:PIRL} in the sense that the agent's behavior is well aligned with the physics principles. Specifically, we focus on 1) the reversibility principle: whether the agent follows the AoA, and 2) the gradient heuristic: whether the agent moves toward the high-SNR direction.  \Cref{fig:aoa-ex}, \ref{fig:aoa-nlos-ex}, and \ref{fig:snr-ex} present the histograms of 1000 sample angle outputs $\hat{\Omega}$ (i.e., moving directions) at a LOS, a 1-NLOS, and a $2^+$-NLOS position, respectively. One can see from these figures that the PIRL obeys the physics principles enforced by $C_{\rm AoA}$ and $C_{\rm SNR}$. 

Furthermore, we attempt to interpret the PIRL model using explainable AI methodologies, such as LIME \cite{riberio12lime}. However, LIME aims to learn an interpretable model (e.g., decision trees) using \textbf{perturbed} training data as a surrogate to the original model. The perturbation is to highlight the features contributing the most to the output. The difficulty of applying LIME in WIN setup is that properly perturbing the wireless field is challenging. Due to diffractions and reflections in mmWave propagation, a slight offset to the target location can create drastically different wireless fields. Hence, as a compromise, we compute the gradient of the PIRL model regarding the input wireless data to inspect whether the instrumental features include AoA in LOS/1-NLOS and SNR in $2^+$-NLOS, as suggested by the physics principles. \Cref{fig:grad-aoa}, \ref{fig:grad-aoa-nlos}, and \ref{fig:grad-snr-ex} empirical confirms that the PIRL model leverages the wireless information as instructed by the principles, which points to another advantage of incorporating the physics information into RL: the physics-based reward components lead to interpretable navigation.

\paragraph{Ablation Study} Recall that PIRL differs from WAN in its use of link state and SNR information. We conduct ablation studies regarding $C_{\rm LS}$ and $C_{\rm SNR}$, for which we report the NPL results.  For the SNR ablation, we replace $C_{\rm SNR}$ with the relative distance cost in $2^+$-NLOS to see whether the SNR heuristic helps the agent navigate efficiently in such a scenario. As one can see from \Cref{tab:ablation}, the answer to the question is affirmative, as the SNR ablation returns significantly higher NPLs in $2^+$-NLOS. We also replace $C_{\rm LS}$ with a constant number to investigate whether the link-state penalty discourages the agent from entering the higher-order NLOS area from the lower-order NLOS. The third row in \Cref{tab:ablation} indicates that without $C_{\rm LS}$, the agent frequently revisits the high-order NLOS areas in testing, which yields higher NPLs in NLOS scenarios. In summary, $C_{\rm SNR}$ contributes to PIRL's success in $2^+$-NLOS, and $C_{\rm LS}$ helps stabilize the navigation (less variance). 
\begin{table}[]
\centering
\caption{Ablation Studies on the SNR and link state terms. The metric is NPL averaged over all testing tasks.}
\label{tab:ablation}
\begin{tabular}{llll}
\toprule
                 & LOS                         & 1-NLOS                      & $2^+$-NLOS                     \\ \hline
WAN                 & $1.01\pm 0.01$ & $1.45\pm 0.03$  & $3.83\pm 0.81$\\ 
PIRL                & $1.01 \pm 0.01$ & $1.41 \pm 0.03$ & $2.60 \pm 0.05$ \\ 
SNR Ablation        & $1.02 \pm 0.02$  & $1.46 \pm 0.04$  & $4.62 \pm 1.15$  \\
Link State Ablation & $1.02 \pm 0.02$ & $1.47 \pm 0.05$  & $3.90 \pm 1.02$  \\ 
\bottomrule
\end{tabular}
\end{table}

\section{Conclusion}
This work develops a physics-informed RL (PIRL) for wireless indoor navigation and is evaluated by the proposed digital twin. By incorporating physics prior into reward shaping, PIRL introduces learning biases to modulate the policy learning favoring those adhering to physics principles. As these principles are invariant across training/testing tasks, PIRL alleviates catastrophic forgetting in training and displays zero-shot generalization in testing.

% \appendices
% \input{algo}

\bibliographystyle{IEEEtran}
\bibliography{reference}

\end{document}